%% file: main.tex
\documentclass[10pt,twocolumn,letterpaper]{article}

\usepackage[final]{cvpr}      

\input{preamble}

\definecolor{cvprblue}{rgb}{0.21,0.49,0.74}
\usepackage[pagebackref,breaklinks,colorlinks,allcolors=cvprblue]{hyperref}

\def\paperID{9770} 
\def\confName{CVPR}
\def\confYear{2025}

\title{Synthetic Data is an Elegant GIFT for Continual Vision-Language Models}

\author{
    Bin Wu$^{1}$\thanks{Equal Contribution. $^{\dagger}$Corresponding Author.} , Wuxuan Shi$^{1}\footnotemark[1]$ , Jinqiao Wang$^{2,3}$ and Mang Ye$^{1,4\dagger}$\\
    $^{1}$ School of Computer Science, Wuhan University, Wuhan, China\\
    $^{2}$ Institute of Automation, Chinese Academy of Sciences, Beijing, China\\
    $^{3}$ Wuhan AI Research, Wuhan, China\\
    $^{4}$ Taikang Center for Life and Medical Sciences, Wuhan University, Wuhan, China\\
    {\tt\small \{wubin2021, wuxuanshi, yemang\}@whu.edu.cn, jqwang@nlpr.ia.ac.cn} \\
    {\tt\small\url{https://github.com/Luo-Jiaming/GIFT_CL}}\\
}

\begin{document}
\maketitle
\input{sec/0_abstract}    
\input{sec/1_intro}
\input{sec/2_relate}
\input{sec/3_method}
\input{sec/4_exps}
\input{sec/5_conclusion}

\input{sec/6_acknowledge}

{
    \small
    \bibliographystyle{ieeenat_fullname}
    \bibliography{main}
}

\input{sec/X_suppl}

\end{document}

%% file: preamble.tex
\usepackage{xcolor}
\usepackage{colortbl}
\usepackage{multirow}

\newcommand{\textBlue}[1]{\textcolor{NavyBlue!80!black}{#1}}
\newcommand{\textRed}[1]{\textcolor{Maroon!80!black}{#1}}
\newcommand{\cellGray}[1]{\cellcolor{gray!20}#1}
\newcommand{\cellPink}[1]{\cellcolor{Thistle!20}#1}

%% file: sec/0_abstract.tex
\begin{abstract}
Pre-trained Vision-Language Models (VLMs) require Continual Learning (CL) to efficiently update their knowledge and adapt to various downstream tasks without retraining from scratch.
However, for VLMs, in addition to the loss of knowledge previously learned from downstream tasks, pre-training knowledge is also corrupted during continual fine-tuning.
This issue is exacerbated by the unavailability of original pre-training data, leaving VLM's generalization ability degrading.
In this paper, we propose GIFT, a novel continual fine-tuning approach that utilizes synthetic data to overcome catastrophic forgetting in VLMs. Taking advantage of recent advances in text-to-image synthesis, we employ a pre-trained diffusion model to recreate both pre-training and learned downstream task data.
In this way, the VLM can revisit previous knowledge through distillation on matching diffusion-generated images and corresponding text prompts.
Leveraging the broad distribution and high alignment between synthetic image-text pairs in VLM's feature space, we propose a contrastive distillation loss along with an image-text alignment constraint.
To further combat in-distribution overfitting and enhance distillation performance with limited amount of generated data, we incorporate adaptive weight consolidation, utilizing Fisher information from these synthetic image-text pairs and achieving a better stability-plasticity balance.
Extensive experiments demonstrate that our method consistently outperforms previous state-of-the-art approaches across various settings.
\end{abstract}

%% file: sec/1_intro.tex
\section{Introduction}
\label{sec:intro}

\begin{figure}[t]
	\centering
	\includegraphics[width=0.85\linewidth]{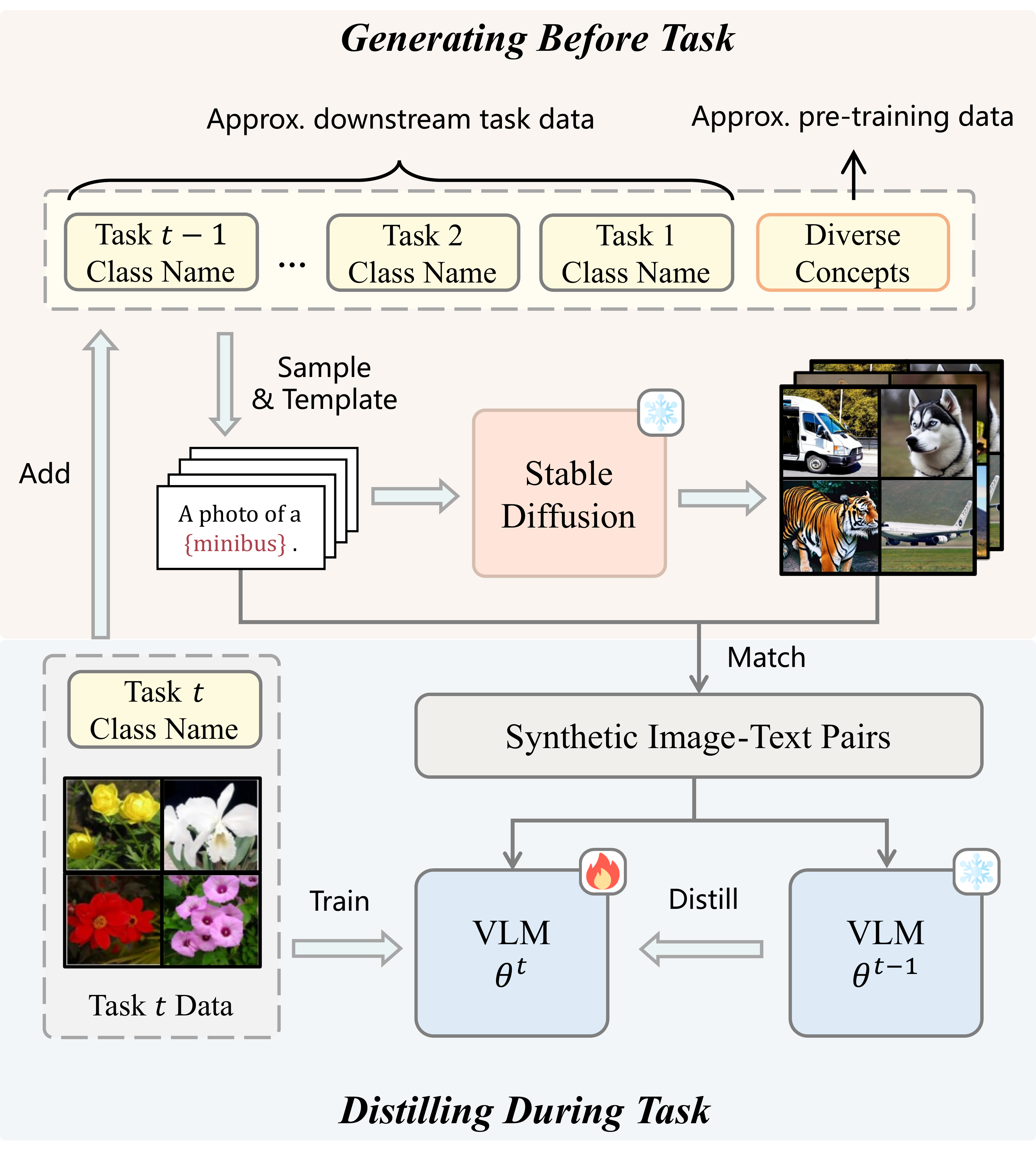}
    \vspace{-1.5ex}
	\caption {
        We use synthetic data generated by Stable Diffusion~\cite{rombach2022high} to support continual fine-tuning of VLMs. By creating prompts from learned downstream class names and diverse visual concepts (i.e., additional class names), the generated data effectively approximates both downstream and VLM’s pre-training data. During training, knowledge distillation~\cite{hinton2015distilling} enables the VLM to re-experience previous knowledge via its past responses on matching synthetic images and corresponding text prompts.
    }
	\label{fig:teaser}
    \vspace{-2.5ex}
\end{figure}

Vision-Language Models (VLMs), such as CLIP~\cite{radford2021learning} and ALIGN~\cite{jia2021scaling}, have offered unprecedented advancements in zero-shot generalization and fine-tuning performance on various downstream tasks. 
Despite these advancements, difficulties remain in updating knowledge of pre-trained VLMs and applying them to multiple downstream tasks simultaneously.
Vanilla approaches involve retraining VLMs from scratch to update knowledge or storing a fine-tuned VLM for each specific task, both of which incur substantial additional costs. 
While Continual Learning (CL)~\cite{kirkpatrick2017overcoming,li2017learning, buzzega2020dark, shi2023prototype, shiprospective} can reduce this cost by enabling trained models to only learn new task data incrementally, thereby presenting itself as an efficient alternative to address these difficulties.

Continual learning faces the fatal problem known as \textit{catastrophic forgetting}~\cite{mccloskey1989catastrophic}, where a model loses previously acquired knowledge upon learning new tasks. 
For continually fine-tuned VLMs, in addition to forgetting learned downstream tasks, forgetting pre-training knowledge significantly impairs their zero-shot generalization ability. 
A straightforward approach to mitigate forgetting is to replay a small portion of stored historical samples~\cite{rebuffi2017icarl,aljundi2019gradient,buzzega2020dark}, but this is impracticable for VLMs due to the inaccessibility of pre-training data.
In parallel, recent advances in text-to-image generative models~\cite{ramesh2021zero,rombach2022high} now enable synthesis of high-quality images from text descriptions, positioning synthetic data as a promising solution for training data scarcity~\cite{he2023synthetic,tian2024stablerep}. Studies~\cite{jodelet2023class,gao2023ddgr,meng2024diffclass} show that images generated by diffusion models~\cite{ho2020denoising,rombach2022high} can effectively approximate historical task data and mitigate forgetting in continual learning of randomly initialized convolutional networks.
This promotes us to ask: \textit{Can synthetic data from latest diffusion models help preserve pre-trained VLMs' knowledge during continual learning, and if so, how?}

To answer this question, we first consider the following two sub-questions:
\textit{(1) How can diffusion model generate to approximate both the pre-training and downstream task data of VLMs?}
\textit{(2) How can the generated data be used to mitigate forgetting?}
For question (1), we prompt a pre-trained Stable Diffusion~\cite{rombach2022high} with class names from downstream tasks in continual learning to generate corresponding synthetic data. Additionally, findings in~\cite{zheng2023preventing} suggest that VLM's pre-training data can be well approximated by semantically rich external datasets like ImageNet~\cite{deng2009imagenet}. 
This insight inspires us to generate pre-training synthetics in the same way as we generate for downstream tasks, but using diverse visual concepts as prompts. These concepts can be sampled from a corpus like the WordNet Synsets~\cite{miller1995wordnet}, or directly from ImageNet class names. As shown in Fig.~\ref{fig:teaser}, we overcome the challenge of inaccessible pre-training data and recreate historical samples without causing privacy and storage issues.
Excessive overhead on image generation is generally undesirable. We achieve domain customization by generating images based on class names, enabling fewer generated images to encompass diverse distributions.
However, the limited generated data volume still carries risks of in-distribution overfitting and reduced effectiveness in forgetting mitigation, necessitating a careful strategy for leveraging the limited synthetic data effectively.
Therefore, as to question (2), we propose a novel continual learning framework for VLMs that leverages \textbf{G}enerated data to \textbf{I}mprove continual \textbf{F}ine-\textbf{T}uning, named \textbf{GIFT}. GIFT comprises two core components: a knowledge distillation~\cite{hinton2015distilling} process to preserve knowledge via synthetic image-text pairs, and an adaptive weight consolidation method for regularization.

Through knowledge distillation, the VLM is encouraged to mimic its original responses for synthetic samples and thus preserve its well-learned feature representations. As a compressed, parameterized image library of diversity, Stable Diffusion shares part of its pre-training data space with the VLM. Its generated images and corresponding text prompts are widely distributed and highly aligned within the VLM's feature space~\cite{hessel2021clipscore}. Based on this, we design a \textbf{contrastive distillation loss} similar to VLM's pre-training objective of image-text matching, as well as an \textbf{image-text alignment constraint} to correct errors accumulated by the teacher model during the distillation process. Furthermore, we find that constraining parameter updates to a close distance from the pre-trained weights improves distillation effectiveness with limited synthetic data. This observation aligns with prior findings~\cite{wortsman2022robust,tian2023trainable} that combining fine-tuning with proper $l_2$ constraints can alleviate in-domain overfitting and maintain out-of-domain robustness, i.e., improving robustness to forgetting. Building on this, we propose \textbf{adaptive weight consolidation} for regularization. Unlike existing weight consolidation methods~\cite{kirkpatrick2017overcoming,zenke2017continual,aljundi2018memory}, our approach dynamically adjusts constraint levels on different parameters based on Fisher information~\cite{pascanu2013revisiting} derived from synthetic data during training, thus preserving model stability without compromising plasticity.
Our contributions are summarized as follows:
\begin{itemize}
    \item We employ stable diffusion to recreate pre-training and historical samples for VLMs without causing privacy and storage issues, which helps the VLMs dynamically recall generic and learned downstream knowledge.
    \item We propose a novel continual fine-tuning framework GIFT that utilizes a small amount of synthetic data to assist VLMs in achieving a better balance between maintaining generalization and learning downstream tasks.
    \item Extensive experiments on 11 datasets across various domains, as well as traditional CL settings, demonstrate that our method consistently outperforms previous state-of-the-art approaches in various scenarios.
\end{itemize}

%% file: sec/2_relate.tex
\section{Related Works}
\label{sec:relate}

\begin{figure*}[t]
	\centering
	\includegraphics[width=1\linewidth]{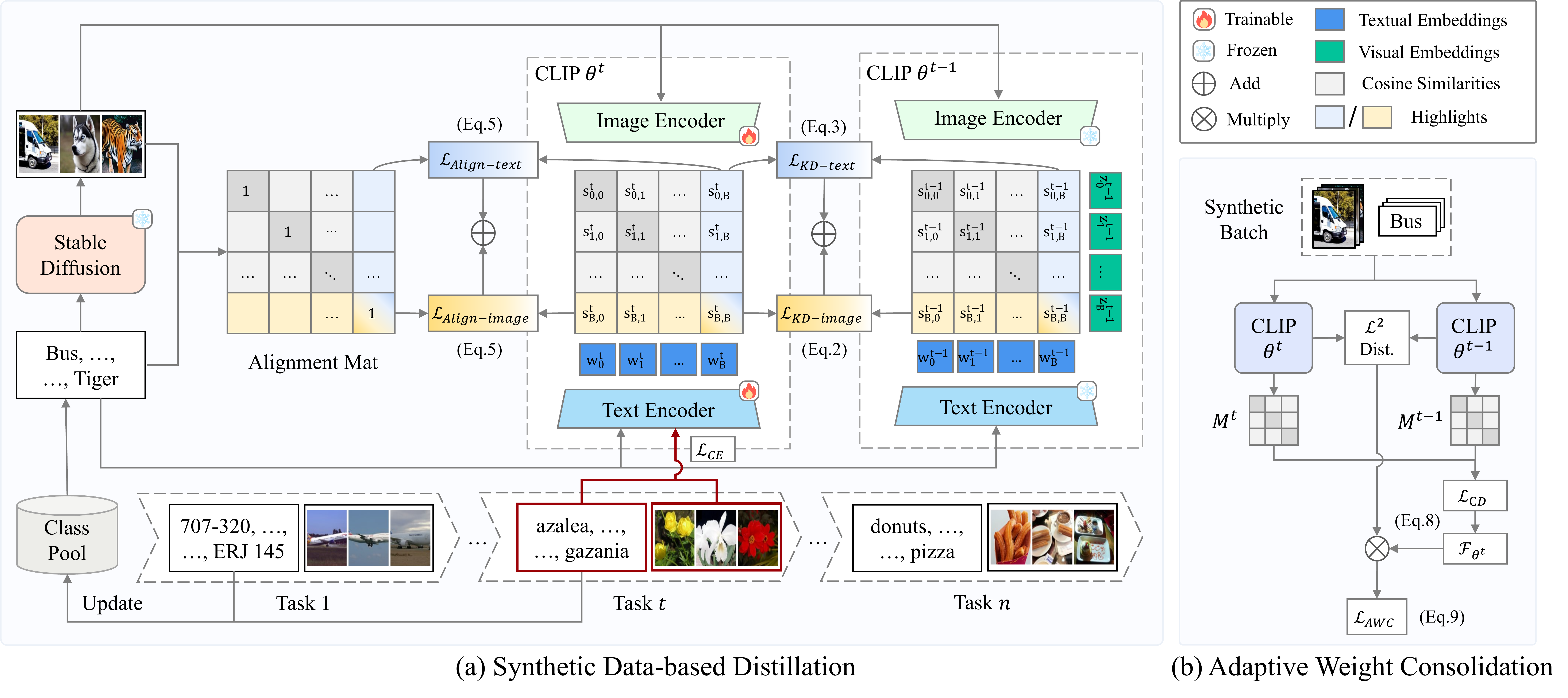}
	\vspace{-4.5ex}
	\caption{Framework overview of GIFT. (a) \textbf{Synthetic Data-based Distillation} aligns the output of the current CLIP model $\theta^t$ with the previous model $\theta^{t-1}$ on matching synthetic image-text pairs when learning a new task. Image-text alignment loss is applied to correct errors in the teacher model through hard target, i.e., the alignment matrix. (b) \textbf{Adaptive Weight Consolidation} employs a parameter importance weighted $l_2$ penalty to limit parameter changes causing forgetting and overfitting. By leveraging the Fisher information $\mathcal{F}_{\theta^t}$ from synthetic image-text pairs during training, parameter importance is adjusted in real-time to achieve a better stability-plasticity balance.}
	\label{fig:frame}
	\vspace{-2.5ex}
\end{figure*}

\noindent\textbf{Continual Learning (CL).}
CL aims to learn from sequential data, incorporating new knowledge without forgetting. 
Most CL approaches can be classified into three categories: architecture-based, regularization-based, and replay-based methods.
Architecture-based methods ~\cite{yoon2017lifelong, wang2022foster,yan2021dynamically,aljundi2017expert,douillard2022dytox} allocate new parameters for new tasks. 
The isolated parameters significantly reduce forgetting but also block knowledge transfer.
Regularization-based methods try to mitigate forgetting by restricting parameter updates based on parameter importance~\cite{kirkpatrick2017overcoming, zenke2017continual, aljundi2018memory}, but limit model plasticity due to static regularization. 
Because these methods fail to adapt estimated parameter importance to the training dynamics, causing outdated and ineffective estimations as the task sequence progresses.
We address this issue by leveraging Fisher information of real-time updated model parameters on synthetic image-text pairs. Meanwhile, knowledge distillation~\cite{hinton2015distilling} is typically used as a regularization term~\cite{li2017learning,douillard2020podnet,ding2022don} to align the current output space with previous ones. Although the effect of knowledge distillation on current task dataset is often insufficient, distillation on stored historical samples achieves better performance when combined with experience rehearsal~\cite{rebuffi2017icarl, buzzega2020dark}. Despite strong performance, replay-based methods~\cite{aljundi2019gradient,rebuffi2017icarl, buzzega2020dark,bang2021rainbow} that store partial historical samples are impractical in scenarios with storage limitations or privacy concerns.

\noindent\textbf{Continual Learning with Synthetic Data.}
Some replay-based methods tackle the issue of inaccessible historical data by leveraging synthetic samples. Early works reproduces historical samples using model inversion~\cite{yin2020dreaming, smith2021always} or generative adversarial network (GAN)~\cite{shin2017continual, he2018exemplar, xiang2019incremental}, but fall short due to high computational costs and low generation quality. Recent works have achieved success in overcoming forgetting with diffusion-generated data in continual learning of convolutional models~\cite{jodelet2023class,gao2023ddgr,meng2024diffclass} and continual object detection~\cite{kim2024sddgr}.
Taking advantage of the strong alignment between feature representations of diffusion-generated images and their text prompts, we introduce diffusion-generated data to the continual learning of VLMs.

\noindent\textbf{Continual Learning for VLMs.}
Pre-training on large-scale image-text pairs endows Vision-Language Models (VLMs)~\cite{li2017learning,jia2021scaling,ye2025survey} with strong generalization and even zero-shot capabilities. 
Extensive research has focused on fine-tuning VLMs to boost downstream task performance while preserving their inherent generalization abilities. 
Several approaches optimize additional structures to avoid altering backbone parameters directly, such as prompt learning~\cite{zhou2022learning, zhou2022conditional,jia2022visual} and adapter tuning~\cite{gao2024clip}. These fine-tuning strategies have been extended to CL settings by introducing task-specific structures~\cite{wang2022learning, wang2022dualprompt, wang2022s, zhou2023learning, yu2024boosting}. However, the limited number of learnable parameters in these methods determines their suboptimal performance on complex tasks. 
Robust fine-tuning of model backbone parameters~\cite{wortsman2022robust,goyal2023finetune,tian2023trainable,huang2024learn} is also explored to achieve a stability-plasticity balance. 
To address continual learning scenarios with domain shift between tasks, ZSCL~\cite{zheng2023preventing} enhances zero-shot capability protection through distillation on a large external reference dataset. We substitute the external reference dataset with a pre-trained generative model, which significantly reduces storage overhead while maintaining the diversity of the distillation data sources.

%% file: sec/3_method.tex
\section{Method}

\subsection{Preliminaries}

\noindent\textbf{Continual Learning.} 
Given a set of $n$ tasks $\{\mathcal{T}^i\}_{i=1}^n$, continual training is conducted sequentially on each task $\mathcal{T}^i=(\mathcal{D}^i, C^i)$. 
Here, $\mathcal{D}^i$ represents the task dataset $\{\mathbf{x}^i_j, y^i_j\}_{j=1}^{N_i}$ with $N_i$ instances, where $\mathbf{x}^i_j$ is an input image belonging to class $y^i_j \in \mathcal{Y}^i$. And $\mathcal{Y}^i$ is the label space of the $i^{th}$ task $\mathcal{T}^i$. Class names $C^i=\{c^i_j\}_{j=1}^{\vert \mathcal{Y}^i \vert}$ maps the image labels to specific object names. The goal of continual learning is to maintain high performance across all tasks. 
We focus on two continual learning settings~\cite{van2019three}. In Task-Incremental Learning (TIL), the image $\mathbf{x}$ to be predicted is provided with its task identity $t$, so the model generates predictions within $C^t$. In Class-Incremental Learning (CIL), the task identity $t$ is not given at inference and the model must distinguish between all the previously encountered classes~$C=\bigcup_{i=1}^tC^i$.

\vspace{1ex}
\noindent\textbf{Vision-Language Model.} 
This paper focuses on Contrastive Language-Image Pre-training (CLIP) ~\cite{radford2021learning} as the VLM. During pre-training, CLIP jointly learns an image encoder $f_i(\cdot)$ and a text encoder $f_t(\cdot)$. Given an input image $\textbf{x}$, the output distribution of CLIP is formulated~as:
\begin{equation} \label{eq:clip_pred}
    p(y_i \mid \mathbf{x})=\frac{\exp \left(\cos \left(\mathbf{z}, \mathbf{w}_i  \right) / \tau\right)}{\sum_{j=1}^{|\mathcal{Y}|} \exp \left(\cos \left(\mathbf{z}, \mathbf{w}_j \right) / \tau\right)} \,,
\end{equation}
where $\cos(\cdot,\cdot)$ denotes cosine similarity, $\tau$ is the temperature parameter learned during pre-training, $\mathbf{z}=f_i(\mathbf{x})$ is the image embedding. Correspondingly, $\mathbf{w}_i$ is the text embedding of class $y_i$ obtained by feeding templated texts, e.g., ``a photo of a \{$c_i$\}.'' into the text encoder. 
We denote the templated text of class $y_i$ as $\mathbf{t}_i$. 
Eq.~\ref{eq:clip_pred} aims to find the most similar text $\mathbf{t}_i$ that maximizes the cosine similarity to the query image. We fully fine-tune the CLIP model for higher performance on downstream tasks, with cross-entropy loss applied to the ground truth and the output distribution.


\subsection{Knowledge Distillation on Synthetic Data}
CLIP's zero-shot generalization ability stems from its well learned feature space that aligns images and texts effectively. 
However, directly fine-tuning on downstream task data distorts this alignment, causing the feature space to collapse into the subspace of specific downstream task and leading to catastrophic forgetting. 
We resurrect pre-training and historical task data using Stable Diffusion and use these synthetic data as anchors to preserve the integrity of the feature space through distillation, thus mitigating forgetting.

\vspace{1ex}
\noindent\textbf{Image Generation.}
While the optimal distillation data sources to preserve zero-shot capability would be the original pre-training dataset, its unavailability necessitates the use of a diverse generated dataset as an effective approximation.
We generate images directly from class names, maintaining a class buffer pool $\mathcal{P}$ that stores all the class names encountered by CLIP in historical downstream tasks, and a base class name set $C^0$.
Before continual learning begins, $\mathcal{P}$ is initialized with $C^0$, a semantically rich class name set used to approximate CLIP's pre-training data.
In particular, we choose the class names of ImageNet~\cite{deng2009imagenet} as $C^0$.
As new downstream tasks are learned, their class names are added to $\mathcal{P}$.
At the start of task $t, t \ge 1$, we randomly sample class names from $\mathcal{P}=\bigcup_{i=0}^tC^i$ with replacement and template them as ``a photo of a \{$c$\}.'' to prompt the pre-trained Stable Diffusion to generate images for distillation.

The similarity between CLIP's and Stable Diffusion's pre-training data allows images generated by Stable Diffusion to closely approximate CLIP's pre-training data distribution \textbf{without fine-tuning}.
Using Stable Diffusion to generate downstream task data also reduces the domain gap between synthetic downstream task data and CLIP's original pre-training data.
Additionally, we ensure inter-class diversity in generated images through the variety of stored class names, resulting in generated images that are widely distributed across CLIP's feature space. 
The low generation cost of Stable Diffusion eliminates the need to store generated images. After completing a task, the images are discarded and regenerated before the next task, further ensuring diversity in distillation data sources.

\vspace{1ex}
\noindent\textbf{Contrastive Distillation.} 
We implement the distillation loss in a contrastive~\cite{goyal2023finetune, ni2023continual} manner to align with the  image-text matching objective of CLIP pre-training.
For a synthetic batch containing $B$ image-text pairs at task $t$, the current model CLIP $\theta^t$ encodes this batch into $l_2$ normalized embeddings $\{(\mathbf{z}_1^t,\mathbf{w}_1^t),(\mathbf{z}_2^t,\mathbf{w}_2^t), \dots, (\mathbf{z}_B^t, \mathbf{w}_B^t)\}$. 
The image-text similarity within the batch is then computed to obtain the contrastive matrix $M^t=[s_{i,j}^t]_{B \times B}$, where $s_{i,j}^t$ represents cosine similarity $\cos(\mathbf{z}_i^t,\mathbf{w}_j^t)$. 
Similarly, we use the previous CLIP $\theta^{t-1}$ from the last task as the teacher model to compute $M^{t-1}$.
The knowledge distillation loss for image classification is then computed using Kullback-Leibler Divergence \cite{csiszar1975divergence} to align $M^{t-1}$ and $M^{t}$ in rows:
\begin{equation} \label{eq:kd_image}
    \mathcal{L}_{KD\_{image}} = -\frac{1}{B}\sum_{i=1}^{B} M^{t-1}_{i,:} \cdot \log \left( \frac{M^{t}_{i,:} }{M^{t-1}_{i,:}} \right)\,,
\end{equation}
where $M^{t-1}_{i,:}$ and $M^{t}_{i,:}$ denote the $i^{th}$ row of $M^{t-1}$ and $M^{t}$ respectively. 
To enhance modality alignment, we compute the text retrieval distillation loss in columns symmetrically:
\begin{equation} \label{eq:kd_text}
    \mathcal{L}_{KD\_{text}} = -\frac{1}{B}\sum_{j=1}^{B} M^{t-1}_{:,j} \cdot \log \left( \frac{M^{t}_{:,j} }{M^{t-1}_{:,j}} \right)\,.
\end{equation}
Symmetrically for both visual and textual modalities, the overall \textbf{C}ontrastive \textbf{D}istillation loss is calculated as:
\begin{equation} \label{eq:kd}
    \mathcal{L}_{CD} = \mathcal{L}_{KD\_{image}} + \mathcal{L}_{KD\_{text}} \,.
\end{equation}

\noindent\textbf{Image-Text Alignment.}
A challenge in knowledge distillation for CL is that the teacher model, having learned one less task than the student model, also suffers from catastrophic forgetting and can make errors on historical tasks. 
These errors propagate during the distillation process, causing a worsening modality mismatch in student model.
To mitigate this, combining distillation soft targets with the ground truth hard targets is a simple yet powerful solution. 
Due to Stable Diffusion pre-training~\cite{rombach2022high}, the generated images exhibit strong alignment with their corresponding textual prompts in CLIP’s feature space. We use this alignment as hard targets to complement the distillation soft targets, correcting image-text mismatches caused by errors of the teacher model and thereby ensuring more reliable knowledge retention.
To implement this image-text alignment constraint, we calculate KL divergence between the $B$-class identity matrix $I_B$ (as the alignment matrix) and the contrastive matrix $M^t$ in a manner similar to Eq.\ref{eq:kd_image} and Eq.\ref{eq:kd_text} respectively, resulting in $\mathcal{L}_{Align\_{image}}$ and $\mathcal{L}_{Align\_{text}}$. Then sum up to get the \textbf{I}mage-\textbf{T}ext \textbf{A}lignment loss $\mathcal{L}_{ITA}$:
\begin{equation} \label{eq:align}
    \mathcal{L}_{ITA} = \mathcal{L}_{Align\_{image}} + \mathcal{L}_{Align\_{text}} \,.
\end{equation}
Together with the \textbf{C}ross \textbf{E}ntropy loss $\mathcal{L}_{CE}$ for learning new task $t$, the total training loss with synthetic data-based distillation $L_{Total}$ can be written as:
\begin{equation} \label{eq:total}
    \mathcal{L}_{Total} = \mathcal{L}_{CE} + \alpha \mathcal{L}_{CD} + \beta \mathcal{L}_{ITA} \,,
\end{equation}
where $\alpha$ and $\beta$ are hyperparameters that balance the trade-off between the terms.

\subsection{Adaptive Weight Consolidation}
VLMs like CLIP tend to overfit to in-distribution data of specific downstream tasks during fine-tuning, which impairs their ability to generalize to out-of-distribution tasks, including previously learned tasks, thereby exacerbating forgetting. 
This is even worse when a limited amount of synthetic data fails to maintain the integrity of the feature space with the weakened distillation effect. 
We introduce adaptive weight consolidation as regularization to alleviate overfitting problems and thus reduce forgetting. To achieve better stability-plasticity balance, we use Fisher information from synthetic image-text pairs during training to adaptively adjust the level of constraints on different parameters.

\vspace{1ex}
\noindent\textbf{Weight Consolidation.} 
In practice, the cross-entropy loss of new task learning tends to dominate the optimization process in the early training stages. Specifically, cross-entropy gradients draw the student model parameters towards a sharp local optimum diverging from the teacher model. This divergence leads to in-distribution overfitting and increased distillation loss, making it difficult to return to the broad minimum where the teacher model resides, even after the distillation loss drops and converges. Weight consolidation~\cite{zenke2017continual} addresses this issue by introducing an $l_2$ penalty, which constrains the model parameters to remain close to robust parameters of the teacher model. EWC~\cite{kirkpatrick2017overcoming}, a typical weight consolidation method, imposes a parameter importance weighted $l_2$ loss as follows:
\begin{equation} \label{eq:ewc}
    \mathcal{L}_{EWC} = \sum_{i} \mathcal{F}_{\theta_{i}^{t-1}} \cdot \left( \theta^{t}_i - \theta^{t-1}_i \right)^2 \,,
\end{equation}
where the parameter importance $\,\smash{\mathcal{F}_{\theta_{i}^{t-1}}}$ is calculated as the diagonal elements of the Fisher Information Matrix~\cite{pascanu2013revisiting}.

\vspace{1ex}
\noindent\textbf{Adapt to Training Dynamics.} 
Although EWC seeks to maintain model plasticity by applying varying levels of constraints on different parameters based on their estimated importance, it falls short due to the static nature of this estimation. EWC calculates Fisher information using the old model on the old task. However, since model parameters change dynamically during new task optimization, the Fisher information quickly becomes outdated, resulting in only blind constraints. To address this limitation, we propose adaptive weight consolidation, which keeps updating the Fisher information throughout the optimization process. As optimization progresses, the knowledge distillation loss serves as a reliable indicator of the model's level of forgetting. Thus, we directly use the distillation loss on the synthetic image-text pairs as the log-likelihood to calculate the diagonal Fisher information:
\begin{equation} \label{eq:fisher}
    \mathcal{F}_{\theta_{i}^t}^{(j)} = \left( \frac{\partial \left( \alpha\mathcal{L}_{KD}^{(j)}+\beta\mathcal{L}_{Align}^{(j)} \right)}{\partial \theta_{i}^t} \right) ^2 \,,
\end{equation}
where $\smash{\mathcal{F}_{\theta_{i}^t}^{(j)}}$ denotes the diagonal Fisher information of model parameter $\theta_{i}^t$ at the $j^{th}$ optimization step. Our adaptive weight consolidation loss is then formulated as:
\begin{equation} \label{eq:awc}
    \mathcal{L}_{AWC}^{(j)} = \sum_{i} \mathcal{F}_{\theta_{i}^{t}}^{(j)}  \cdot \left( \theta^{t(j)}_i - \theta^{t-1}_i \right)^2 \,.
\end{equation}

Notably, $\smash{\mathcal{F}_{\theta_{i}^t}^{\left( j\right)}}$ is also the squared gradients of the distillation loss, reflecting its stability. When learning new tasks that are significantly different from those previously learned by CLIP, most of the gradient directions of $\mathcal{L}_{CE}$ oppose those of distillation loss. 
In such cases, $\mathcal{L}_{AWC}$ can constrain parameter updates that are likely to cause drastic changes in distillation loss, i.e., parameter updates that aggravate overfitting and forgetting. This helps smooth out conflicts among multiple optimization objectives and stabilize the distillation loss without compromising plasticity. Our adaptive update of Fisher information utilizes the intermediate results from the back propagation of the distillation loss and introduces minimal computational overhead.

%% file: sec/4_exps.tex
\section{Experiments}
\subsection{Experimental Setting}
\noindent\textbf{Datasets.}
We evaluate our method on two settings: Multi-domain TIL (MTIL) and CIL. MTIL is specifically designed for CL of VLMs and presents a significant challenge as it encompasses 11 datasets with a total of 1,201 classes across various domains, including Aircraft~\cite{maji2013fine}, Caltech101~\cite{fei2004learning}, CIFAR100~\cite{krizhevsky2009learning}, DTD~\cite{cimpoi2014describing}, EuroSAT~\cite{helber2019eurosat}, Flowers~\cite{nilsback2008automated}, Food~\cite{bossard2014food}, MNIST~\cite{deng2012mnist}, OxfordPet~\cite{parkhi2012cats}, StanfordCars~\cite{krause20133d}, and SUN397~\cite{xiao2010sun}.
We follow the two-order training protocol proposed in~\cite{zheng2023preventing} for MTIL. 
The ablation experiments are conducted on MTIL order I by default.
Experiments on the CIL setup are presented in the supplementary material.

\vspace{1ex}
\noindent\textbf{Metrics.}
To evaluate our method on the MTIL setting, we utilize metrics proposed in~\cite{zheng2023preventing}, namely ``Transfer'', ``Avg.'', and ``Last''. The ``Transfer'' metric assesses the model's zero-shot capability on unseen data within the task sequence and further reflects the forgetting of pre-training knowledge. ``Last'' evaluates the model's ability to retain historical downstream knowledge. ``Avg.'' is a composite metric assesses the mean performance across ``Transfer" and ``Last", reflecting the stability-plasticity balance. 

\vspace{0.5ex}
\noindent\textbf{Implementation Details.}
As in~\cite{zheng2023preventing}, we use the CLIP model with ViT-B/16~\cite{dosovitskiy2020image} as our backbone for all experiments. We train 1K iterations with batch size 64 for each task in MTIL. The introduced hyperparameters $\alpha$ and $\beta$ are set to 1 and 0.25 respectively to ensure stable results.  For image synthesis, we use Stable Diffusion v1.5~\cite{rombach2022high} with classifier-free guidance set to 7.5 and 50 denoising steps, generating 1K synthetic images per task. More implementation details can be found in the supplementary material.

\subsection{Comparison with State-of-the-art Methods}

\begin{table}[t]
	\centering
    \caption{Comparison of SOTA methods on MTIL Order I.}
    \vspace{-1.5ex}
	\resizebox{1\columnwidth}{!}{%
		\begin{tabular}{@{}l|cc|cc|cc@{}}
			\toprule
			Method & Transfer & $\Delta$ & Avg. & $\Delta$ & Last & $\Delta$  \\ 
            \midrule
            Zero-shot  & 69.4 &  - & 65.3 & - & 65.3 & - \\
            Continual Finetune & 44.6 & - & 55.9 & - & 77.3 & - \\
            \midrule
            \rowcolor{gray!20}
            $l_2$ baseline & 61.0 & 0.0 & 62.7 & 0.0 & 75.9 & 0.0 \\
            \midrule
            LwF~\cite{li2017learning} & 56.9 & \textBlue{-4.1} & 64.7 & \textRed{+2.0} & 74.6 & \textBlue{-1.3} \\
            iCaRL~\cite{rebuffi2017icarl} & 50.4 & \textBlue{-10.6} & 65.7 & \textRed{+3.0} & 80.1 & \textRed{+4.2} \\
            LwF-VR~\cite{ding2022don} & 57.2 & \textBlue{-3.8} & 65.1 & \textRed{+2.4} & 76.6 & \textRed{+0.7}  \\
            WiSE-FT~\cite{wortsman2022robust} & 52.3 & \textBlue{-8.7} & 60.7 & \textBlue{-2.0} & 77.7 & \textRed{+1.8} \\
            ZSCL~\cite{zheng2023preventing} & 68.1 & \textRed{+7.1} & 75.4 & \textRed{+12.7} & 83.6 & \textRed{+7.7} \\
            MoE-Adapter~\cite{yu2024boosting} & 68.9 & \textRed{+7.9} & 76.7 & \textRed{+14.0} & 85.0 & \textRed{+9.1} \\
            \midrule
            \rowcolor{Thistle!20}
            GIFT (Ours) & \textbf{69.3} & \textbf{\textRed{+8.3}} & \textbf{77.3} & \textbf{\textRed{+14.6}} & \textbf{86.0} & \textbf{\textRed{+10.1}} \\
            \bottomrule
		\end{tabular}%
	}
    \vspace{-1ex}
	\label{tab:result_o1}
\end{table}

\begin{table}[t]
    \centering
    \caption{Comparison of SOTA methods on MTIL Order II.}
    \vspace{-1.5ex}
    \resizebox{1\columnwidth}{!}{%
        \begin{tabular}{@{}l|cc|cc|cc@{}}
            \toprule
            Method & Transfer & $\Delta$ & Avg. & $\Delta$ & Last & $\Delta$  \\ 
            \midrule
            Zero-shot & 65.4 & - & 65.3 & - & 65.3 & - \\
            Continual Finetune & 46.6 & - & 56.2 & - & 67.4 & - \\
            \midrule
            \rowcolor{gray!20}
            $l_2$ baseline & 60.6 & 0.0 & 68.8 & 0.0 & 77.2 & 0.0 \\
            \midrule
            LwF~\cite{li2017learning} & 53.2 & \textBlue{-7.4} & 62.2 & \textBlue{-6.6} & 71.9 & \textBlue{-5.3} \\
            iCaRL~\cite{rebuffi2017icarl} & 50.9 & \textBlue{-9.7} & 56.9 & \textBlue{-11.9} & 71.6 & \textBlue{-5.6} \\
            LwF-VR~\cite{ding2022don} & 53.1 & \textBlue{-7.5} & 60.6 & \textBlue{-8.2} & 68.3 & \textBlue{-3.9} \\
            WiSE-FT~\cite{wortsman2022robust} & 51.0 & \textBlue{-9.6} & 61.5 & \textBlue{-7.3} & 72.2 & \textBlue{-5.0} \\
            ZSCL~\cite{zheng2023preventing} & 64.2 & \textRed{+3.6} & 74.5 & \textRed{+5.7} & 83.4 & \textRed{+6.2} \\
            MoE-Adapter~\cite{yu2024boosting} & 64.3 & \textRed{+3.7} & 74.7 & \textRed{+5.9} & 84.1 & \textRed{+6.9} \\
            \midrule
            \rowcolor{Thistle!20}
            GIFT (Ours) & \textbf{65.9} & \textbf{\textRed{+5.3}} & \textbf{75.7} & \textbf{\textRed{+6.9}} & \textbf{85.3} & \textbf{\textRed{+8.1}} \\
            \bottomrule
        \end{tabular}%
    }
    \label{tab:result_o2}
    \vspace{-2ex}
\end{table}

Tab.~\ref{tab:result_o1} and Tab.~\ref{tab:result_o2} display the performance of different methods on the MTIL benchmark, in order I and order II respectively (detailed in the supplementary material). The different arrangements in order I and order II introduce varying degrees of domain shift. 
In the Method column, ``Zero-shot'' denotes the zero-shot performance of the initial CLIP model, indicating the upper bound of the Transfer metric. ``Continual Finetune'' refers to continual fine-tuning without any protection, suggesting the lower bound of Avg. and Last metrics. We also introduce a straightforward baseline, referred to as the ``$l_2$ baseline'', which employs a direct, unweighted $l_2$ penalty to prevent significant changes in model parameters before and after learning a new task. Previous methods~\cite{li2017learning,rebuffi2017icarl,ding2022don,wortsman2022robust} have shown limited improvement in the Last metric, and their Transfer metric performance is significantly lower compared to the $l_2$ baseline, as most of them don't take into account the forgetting of pre-training knowledge. We attribute the non-trivial performance of the $l_2$ baseline, particularly in MTIL order II with its significant domain shift, to the fact that the optimum reached by CLIP through pre-training is broad and flat. Our proposed GIFT outperforms all other methods across all metrics, including the strong competitors ZSCL~\cite{zheng2023preventing} and MoE-Adapter~\cite{yu2024boosting}. 
MoE-Adapter introduces extra parameters that grow linearly but performs worse when the domain shift is drastic (i.e., in MTIL Order II). It's worth nothing that our method using 1K synthetic images outperforms ZSCL using 100K ImageNet images, demonstrating the effectiveness of synthetic data in balancing between preserving general knowledge and retaining downstream task learning.

\begin{table}[t]
    \centering
    \caption{Ablation study of different components.}
    \vspace{-1.5ex}
    \resizebox{1.0\columnwidth}{!}{%
        \begin{tabular}{@{}ccc|cc|cc|cc@{}}
            \toprule
            \multicolumn{3}{l|}{Method} & Transfer & $\Delta$ & Avg. & $\Delta$ & Last & $\Delta$  \\ 
            \midrule
            \multicolumn{3}{l|}{Zero-shot} & 69.4 & - & 65.3 & - & 65.3 & - \\
            \multicolumn{3}{l|}{Continual Finetune} & 44.6 & - & 55.9 & - & 77.3 & - \\
            \midrule
            \multicolumn{3}{l|}{\cellGray{$l_2$ Baseline}} & \cellGray{61.0} & \cellGray{0.0} & \cellGray{62.7} & \cellGray{0.0} & \cellGray{75.9} & \cellGray{0.0} \\
            \hline
            \midrule
            +CD & +ITA & +AWC & Transfer & $\Delta$ & Avg. & $\Delta$ & Last & $\Delta$  \\ 
            \midrule
            & $\surd$ & & 63.5 & \textRed{+2.5} & 70.5 & \textRed{+7.8} & 78.6 & \textRed{+2.7} \\
            $\surd$ & & & 68.3 & \textRed{+7.3} & 76.3 & \textRed{+13.6} & 84.7 & \textRed{+8.8} \\
            $\surd$ & $\surd$ & & 68.9 & \textRed{+7.9} & 76.6 & \textRed{+13.9} & 85.0 & \textRed{+9.1} \\
            $\surd$ & & $\surd$ & 68.7 & \textRed{+7.7} & 77.0 & \textRed{+14.3} & 85.8 & \textRed{+9.9} \\
            $\cellPink{\surd}$ & $\cellPink{\surd}$ & $\cellPink{\surd}$ & \cellPink{\textbf{69.3}} & \cellPink{\textbf{\textRed{+8.3}}} & \cellPink{\textbf{77.3}} & \cellPink{\textbf{\textRed{+14.6}}} & \cellPink{\textbf{86.0}} & \cellPink{\textbf{\textRed{+10.1}}} \\
            \bottomrule
        \end{tabular}%
    }
    \vspace{-2.5ex}
    \label{tab:ablation}
\end{table}

\begin{table*}[t]
    \centering
    \caption{\textbf{Analysis of distillation mechanism.} The default settings are marked in \colorbox{gray!20}{gray}, which employs a contrastive distillation loss, the last CLIP model as the teacher model, and $\beta=0.25$ for ITA scale.}
    \vspace{-1ex}
    \resizebox{0.9\linewidth}{!}{%
        \begin{minipage}{0.32\linewidth} 
            \centering
            \caption*{(a) \textbf{Distillation Loss.}}
            \vspace{-2ex}
            \begin{tabular}{@{}l|c|c|c@{}}
                \toprule
                Loss & Transfer & Avg. & Last  \\
                \midrule
                Feat. Dist. &64.0&71.6&80.5\\
                Image-only &66.8&75.1&84.1\\
                Text-only &64.7&71.9&81.8\\
                \rowcolor{gray!20}
                Contrastive & \textbf{68.9} & \textbf{76.6} & \textbf{85.0} \\
                \bottomrule
            \end{tabular}
        \end{minipage}%
        \hspace{0.05\linewidth} 
        \begin{minipage}{0.32\linewidth} 
            \centering
            \caption*{(b) \textbf{Teacher Model.}}
            \vspace{-2ex}
            \begin{tabular}{@{}l|c|c|c@{}}
                \toprule
                Teacher & Transfer & Avg. & Last  \\
                \midrule
                Initial CLIP  &69.1 &74.0 &80.1 \\
                \rowcolor{gray!20}
                Last CLIP  & 68.9 & \textbf{76.6} & \textbf{85.0} \\
                WiSE(0.2) &69.1  &76.1  &83.4 \\
                WiSE(0.5) &\textbf{69.6}&75.3&81.6\\
                \bottomrule
            \end{tabular}
        \end{minipage}%
        \hspace{0.05\linewidth} 
         \begin{minipage}{0.32\linewidth} 
            \centering
            \caption*{(c) \textbf{Scale of Image-Text Alignment.}}
            \vspace{-2ex}
            \begin{tabular}{@{}l|c|c|c@{}}
                \toprule
                ITA Scale & Transfer & Avg. & Last  \\
                \midrule
                $\beta=0.0$ &68.3&76.3&84.7 \\
                \rowcolor{gray!20}
                $\beta=0.25$ & \textbf{68.9} & \textbf{76.6} & \textbf{85.0}\\
                $\beta=0.5$ &68.7&76.2&84.2 \\
                $\beta=1.0$ &68.5&75.4&82.4 \\
                \bottomrule
            \end{tabular}
        \end{minipage}
    }
    \label{tab:ablation_GD}
    \vspace{-2ex}
\end{table*}

\subsection{Ablation Study of Different Components}
In this section, we perform ablation studies on our proposed method. We decompose our approach into three components: Contrastive Distillation (CD), Image-Text Alignment (ITA), and Adaptive Weight Consolidation (AWC). CD and ITA together constitute the synthetic data-based distillation. Each component is incrementally added to the $l_2$ baseline, and the experimental results are summarized in Tab.~\ref{tab:ablation}. The performance of using only ITA is significantly worse than using only CD when both are scaled to 1. This underscores the advantage of distillation's soft targets over complete hard targets in terms of memory retention. Soft targets encapsulate more latent information that can reflect learned knowledge, and thus is more robust to the possible noise in synthetic data. When the scales of CD and ITA are set to 1 and 0.25 respectively, the Transfer performance improves by 0.6\%. This suggests that, in the right proportion, hard targets can correct erroneous information in soft targets provided by the teacher model, ensuring accurate knowledge retention. Finally, as an enhanced version of $l_2$ weight consolidation, AWC is employed to dynamically adjust the constraint level during training. This adaptive approach enhances both the stability and plasticity of the model, leading to overall improvements across all three metrics.

\subsection{Analysis of Synthetic Data-based Distillation}
\label{sec:Analysis of Distillation Mechanism}

In this section, we conduct ablation studies and analysis on the synthetic data-based distillation. 
To isolate the effect of AWC, we apply CD and ITA to the $l_2$ baseline. 
The analysis primarily focuses on three aspects of distillation (objective function, teacher model and data source), along with the scale of our proposed complementary hard target ITA.

\noindent\textbf{Distillation Loss.} Tab.~\ref{tab:ablation_GD} (a) compares several types of distillation loss. The feature distance loss \cite{zhu2021prototype}, penalizes $l_2$ distance of the visual and textual embeddings between the teacher and student models separately, performs poorly due to lack of consideration for modality alignment. We also implement image-only and text-only distillation by excluding one of the two encoders from the loss computation in the student model, as introduced in \cite{zheng2023preventing}. Since both encoders suffer from forgetting, these two underperform compared to our contrastive distillation loss. 

\noindent\textbf{Teacher Model.} Tab.~\ref{tab:ablation_GD} (b) shows the distillation performance with different teacher models. Utilizing an initial CLIP that hasn't been trained on downstream tasks yields higher Transfer values. However, emulating the output of such a model weakens the memory of downstream tasks. Using WiSE-FT~\cite{wortsman2022robust} to linearly interpolate between the initial CLIP and the previous CLIP also enhances Transfer performance but still fails to achieve a good balance with learning new tasks and retaining historical tasks.

\noindent\textbf{Scale of Image-Text Alignment. } Tab.~\ref{tab:ablation_GD} (c) illustrates the distillation performance across varying scales of ITA, with the $\alpha$ value fixed to 1. A smaller $\beta$ balances the soft and hard targets well and corrects errors in the soft targets. While a larger $\beta$ can cause the model to overfit the synthetic samples, impeding model's ability to learn new tasks.

\begin{figure}[t]
    \centering
    \includegraphics[width=0.9\columnwidth]{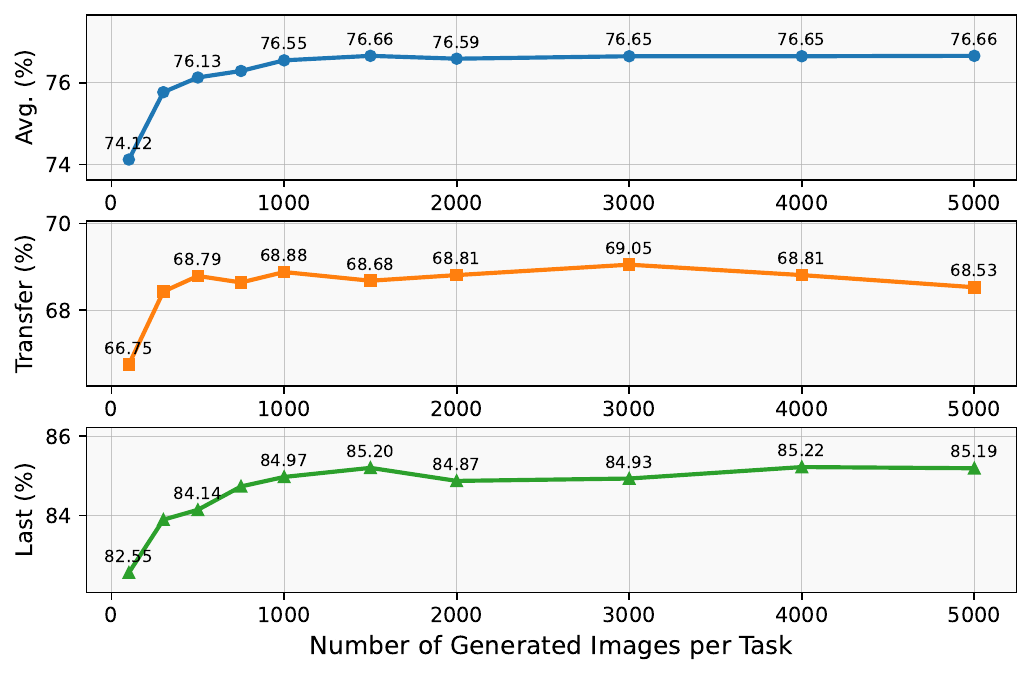}
    \vspace{-2ex}
    \caption{Generating different numbers of synthetic images as distillation data sources for each task produces different results.}
    \vspace{-2ex}
    \label{fig:image_num}
\end{figure}

\noindent\textbf{Number of Synthetic Images.} Fig.~\ref{fig:image_num} presents the ablation study on the number of synthetic images generated for distillation per task. 
To reduce instability caused by varying generated images, we use a consistent seed setting, ensuring smaller image sets are subsets of larger ones. The results show that increasing the number of synthetic images broadens the distribution of anchor points in CLIP's feature space, thereby improving distillation effect. However, performance gains plateau beyond a certain scale (at 1K images). 
Therefore, we conclude that generating 1000 images is both an economical and effective choice.

\begin{table}[t]
    \centering
    \caption{\textbf{Comparison of data sources for distillation.} ``Synthetic'' represents the default setting, which uses both synthetic pre-training and synthetic downstream task data. ``Synthetic$\dagger$'' uses only synthetic pre-training data, i.e., a synthetic ImageNet.}
    \vspace{-1ex}
    \resizebox{0.95\columnwidth}{!}{%
    \begin{tabular}{@{}l|l|c|c|c|c@{}}
        \toprule
        Method & \multicolumn{1}{c|}{Source} & Num & Transfer & Avg. & Last  \\
        \midrule
        \multirow{3}{*}{GIFT} & ImageNet & \multirow{3}{*}{1K} & 69.0 & 76.0 & 84.0 \\
                              & Synthetic &                    & 68.9 \textBlue{(-0.1)} & 76.6 \textRed{(+0.6)} & 85.0 \textRed{(+1.0)} \\
                              & Synthetic$\dagger$ &        & 68.3 \textBlue{(-0.7)} & 75.5 \textBlue{(-0.5)} & 83.7 \textBlue{(-0.3)} \\
        \midrule
        \multirow{3}{*}{GIFT} & ImageNet & \multirow{3}{*}{3K} & 69.3 & 76.6 & 84.7 \\
                              & Synthetic &                    & 69.1 \textBlue{(-0.2)} & 76.7 \textRed{(+0.1)} & 84.9 \textRed{(+0.2)} \\
                              & Synthetic$\dagger$ &        & 69.0 \textBlue{(-0.3)} & 75.9 \textBlue{(-0.7)} & 84.1 \textBlue{(-0.6)} \\
        \midrule
        ZSCL & ImageNet & 100K & 68.1 & 75.4 & 83.6 \\
        \bottomrule
    \end{tabular}%
    }
    \label{tab:image_source}
    \vspace{-2.5ex}
\end{table}

\noindent\textbf{Comparison with Real Images.} Tab.~\ref{tab:image_source} presents the experimental results of replacing synthetic images with an equivalent number of real images sampled from ImageNet~\cite{deng2009imagenet} for each task. With a limited allowance of 1K images, using synthetic images yields higher Avg. and Last values compared to ImageNet images, while maintaining a comparable Transfer value. Because ImageNet images are closer to CLIP's pre-training data and differ in domain from certain downstream tasks, replaying ImageNet data exacerbates forgetting of these tasks, resulting in a lower Last value despite the higher Transfer score. Using synthetic images offers the advantage of balancing the retention of both pre-training and downstream task knowledge. To illustrate this, we generate data solely based on ImageNet labels (``Synthetic$\dagger$''), resulting in performance significantly lower. This outcome also reflects the issue of greater noise in synthetic images compared to real ImageNet data.

\begin{table}[t!]
    \centering
    \caption{\textbf{Comparison of AWC and EWC variant.} ``EWC-V'' refers to the variant of EWC which uses static parameter importance estimation based on synthetic data.}
    \vspace{-1ex}
    \resizebox{0.75\columnwidth}{!}{%
    \begin{tabular}{@{}l|c|c|c|c@{}}
        \toprule
        Method & Num of Samples & Transfer & Avg. & Last  \\
        \midrule
        EWC-V & 64 & 65.8 &75.6 &85.6 \\
        EWC-V & 128 &66.8 & 76.1 & 85.9 \\
        EWC-V & 256 &68.0 &76.8 & \textbf{86.2} \\
        \midrule
        \rowcolor{Thistle!20}
        AWC & -& \textbf{69.3} & \textbf{77.3} & 86.0 \\
        \bottomrule
    \end{tabular}%
    }
    \label{tab:ewc}
    \vspace{-2.5ex}
\end{table}

\subsection{Analysis of Adaptive Weight Consolidation}

\begin{figure}[t]
    \centering
    \vspace{-0.5ex}
    \includegraphics[width=0.8\columnwidth]{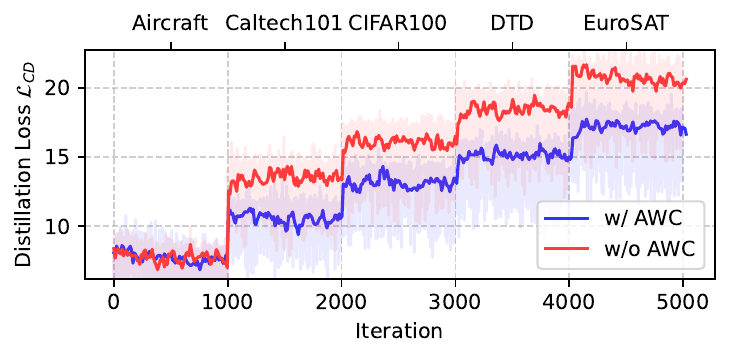}
    \vspace{-2ex}
    \caption{\textbf{CD loss for the first 5 tasks in MTIL order I.} In our implementation, cross-entropy is used as equivalent instead of KL divergence to compute $\mathcal{L}_{CD}$, and the results are presented accordingly. Lower loss means better mitigation of forgetting.}
    \vspace{-1.5ex}
    \label{fig:loss}
\end{figure}

\begin{figure}[t!]
    \centering
    \includegraphics[width=1.0\columnwidth]{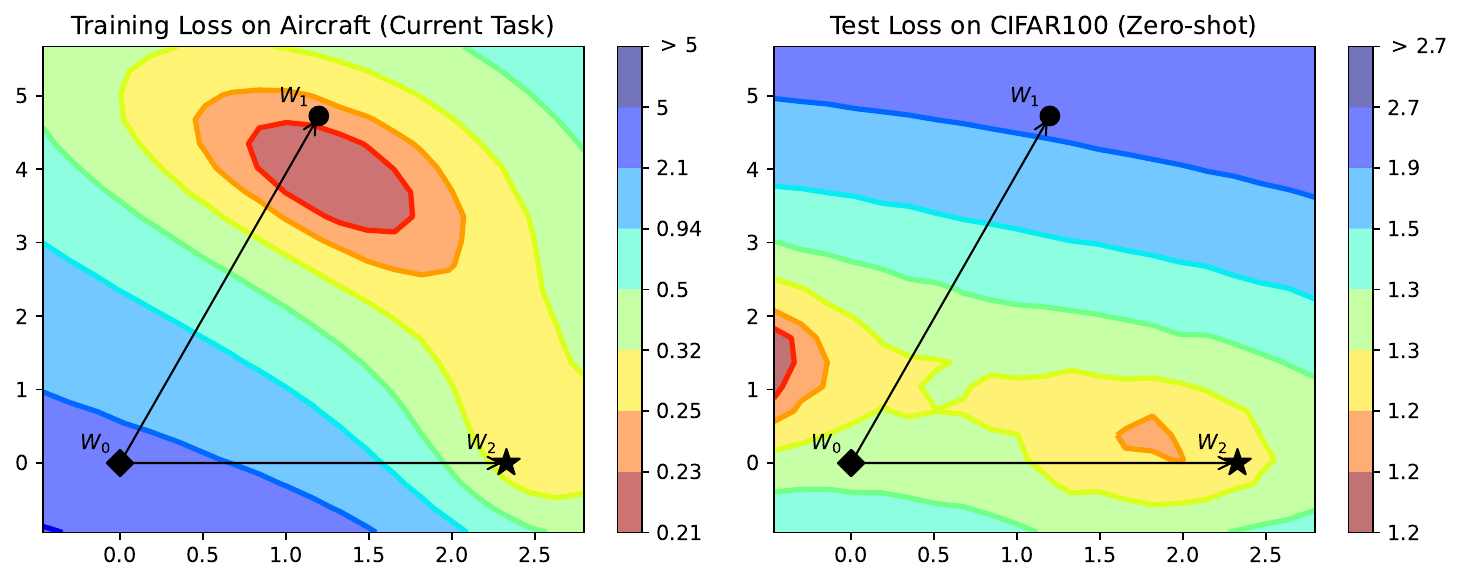}
    \vspace{-4ex}
    \caption{\textbf{Loss values on a two dimensional slice of the loss landscapes.} We use $W_0$, $W_1$ and $W_2$ to represent the initial CLIP and CLIP models finetuned on the Aircraft \cite{maji2013fine} dataset without and with AWC, respectively. As in \cite{garipov2018loss}, we obtain an orthonormal basis $u_1$, $u_2$ for the plane spanned by these models, and the x and y-axis show movement in parameter space in these two directions.}
    \label{fig:surface}
    \vspace{-3ex}
\end{figure}

We ablate and analyze our adaptive weight consolidation method on synthetic data-based distillation without $l_2$ constraints. The results are shown in Fig.~\ref{fig:loss}, Fig.~\ref{fig:surface} and Tab.~\ref{tab:ewc}.

\vspace{0.5ex}
\noindent\textbf{Effect on Distillation Loss.} To better represent the distillation loss, we visualize the cross-entropy between the output distributions of the teacher and student models instead of the KL divergence in Fig.~\ref{fig:loss}, i.e., $H(M^{t-1},M^t) = D_{KL}(M^{t-1} \parallel M^t) + H(M^{t-1})$. The reason is that KL divergence primarily measures the alignment with the teacher model's distribution, which may not accurately reflect the effectiveness of distillation in mitigating forgetting when the teacher model is suboptimal. In Figure \ref{fig:loss}, a higher initial distillation loss at every 1Kth iteration indicate a poorer teacher model. By incorporating AWC, we can achieve a better teacher model and improved distillation performance.

\vspace{0.5ex}
\noindent\textbf{Effect on Combating Overfitting.} Fig.~\ref{fig:surface} illustrates the impact of AWC on combating overfitting by visualizing the train loss on the Aircraft dataset and the test loss on unlearned CIFAR100. Fine-tuning the initial CLIP model on Aircraft, without and with AWC, yields two distinct local optima $W_1$ and $W_2$, where $W_2$ is flatter and broader than $W_1$. Meanwhile, for the test loss on CIFAR100, which reflects the model's zero-shot capability, $W_1$ falls into a high-loss region, whereas $W_2 $ remains within the broad minimum $W_0$ of the initial CLIP model. AWC guides the optimization process to find the overlapping area between the current and historical optimization target within broad minimum of initial CLIP, avoiding falling into the sharp local minimum trap and thus alleviating overfitting.

\vspace{0.5ex}
\noindent\textbf{Comparison with EWC.} To verify the effectiveness of AWC in dynamically estimating and updating parameter importance, we implement a variant method similar to EWC \cite{kirkpatrick2017overcoming} that uses static parameter importance estimation. In its original design, EWC offers no protection during the first task, which poses a significant disadvantage for VLMs with pre-training knowledge. We develop an EWC variant that extracts several batches of synthetic data before each task begins to compute Fisher information as a static parameter importance weight. This weight remains fixed throughout the training process of a downstream task. As shown in Tab.~\ref{tab:ewc}, increasing the number of samples used to compute the static parameter importance improves the performance of this EWC variant. However, it still falls short of AWC, which better accounts for the training dynamics.


%% file: sec/5_conclusion.tex
\section{Conclusion}
In this paper, we propose GIFT, a novel continual fine-tuning approach for VLMs that utilizes diffusion-generated data to mitigate forgetting. GIFT includes a distillation process that encourages the VLM to revisit learned knowledge on matching synthetic image-text pairs and an adaptive weight consolidation strategy for regularization. Since the introduction of synthetic data recreates both pre-training and downstream data without storage and privacy issues, we achieve a balance between preserving general knowledge and retaining downstream task learning in VLMs. Furthermore, our results show that distillation with $l_2$ constraints yields improved performance, allowing stable results with less synthetic data generated. Fisher information from synthetic data enables adaptive $l_2$ constraint, achieving a better stability-plasticity balance.
Comprehensive experiments exhibit superior performance compared to the SOTA.

%% file: sec/6_acknowledge.tex
\vspace{0.5ex}
\noindent\textbf{Acknowledgements.} This work is partially supported by the National Key Research and Development Program of China (2024YFC3308400), National Natural Science Foundation of China under Grant (62176188, 62361166629).

%% file: sec/X_suppl.tex
\clearpage
\setcounter{page}{1}
\maketitlesupplementary

\section*{A. Additional Implementation Details}
We use a batch size 64 for both the MTIL and CIL benchmarks. We employ AdamW~\cite{loshchilov2017decoupled} optimizer ($\beta_1=0.9$, $\beta_2=0.999$) and a label smoothing~\cite{muller2019does} technique for better results. Label smoothing can substitute the regularization of weight decay and achieve better performance. For MTIL and CIFAR100~\cite{krizhevsky2009learning} of CIL, label smoothing is set to 0.2, with weight decay at 0. For TinyImageNet~\cite{le2015tiny} of CIL, we set label smoothing to 0 and weight decay to 0.1. 
Learning rates are searched among $\{10^{-5}, 10^{-6}, 10^{-7}\}$ with cosine annealing~\cite{loshchilov2016sgdr}. 
For the base class set $C^0$ used to prompt Stable Diffusion to generate approximate pre-training data of VLMs, we select all 1000 ImageNet~\cite{deng2009imagenet} class names for MTIL and 100 ImageNet class names that do not overlap with downstream datasets for CIL.

\section*{B. Results on CIL Benchmarks}
\noindent\textbf{Benchmark Description.}
For the CIL setting, we conduct experiments on CIFAR100~\cite{krizhevsky2009learning} and TinyImageNet~\cite{le2015tiny} datasets following~\cite{douillard2022dytox}. 
The 100 classes of CIFAR100 are divided into subsets of $\{10, 20, 50\}$, while the 100 classes from TinyImageNet are divided into subsets of $\{ 5, 10, 20\}$ to evaluate class distribution adaptability. 
For metrics, we adhered to the evaluation protocol in~\cite{douillard2022dytox}, calculating the average accuracy across all datasets at all timestamps (``Avg.'') and the average performance of all tasks after continual learning (``Last'').

\vspace{1ex}
\noindent\textbf{Compared Methods.}
We compare our method with state-of-the-art approaches in the CIL setting (methods listed above ``CLIP Zero-shot'' in Tab.~\ref{tab:cifar100_cil} and Tab.~\ref{tab:tiny_imagenet_cil}). The backbone used by these methods is consistent with that in the papers where they are proposed, i.e., ViT~\cite{dosovitskiy2020image} or Res-Net~\cite{he2016deep}. Like MTIL, we also implement LwF~\cite{li2017learning}, iCaRL~\cite{rebuffi2017icarl}, Lwf-VR~\cite{ding2022don}, and ZSCL~\cite{zheng2023preventing} with CLIP as the backbone and include them in the comparison. The results of CLIP zero-shot predictions and continual fine-tuning without protection are also included, denoted as ``CLIP Zero-shot'' and ``CLIP Fine-tune'', respectively.

\vspace{1ex}
\noindent\textbf{Result Analysis.} As summarized in Tab.~\ref{tab:cifar100_cil} and Tab.~\ref{tab:tiny_imagenet_cil}, experimental results on two CIL settings demonstrate that our method remains effective in single-domain continual learning. Although CLIP excels in zero-shot prediction under these conditions, some CLIP-based continual learning methods are outperformed by those utilizing alternative backbones. This disparity is primarily due to the more pronounced in-distribution overfitting and forgetting that occur with smaller task step sizes. Despite these challenges, our method maintains strong performance even when the number of tasks is large and the incremental step size is small.

\section*{C. Detailed Results on MTIL Benchmark}
\begin{figure*}[h!]
  \centering
  \begin{minipage}{1\textwidth}
    \centering
    \includegraphics[width=1.0\textwidth]{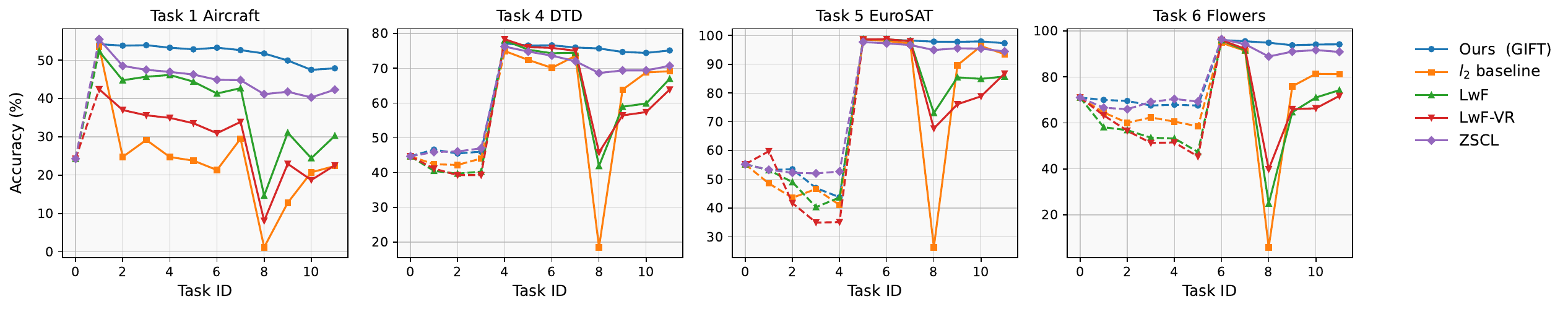}
    \vspace{-5ex}
    \caption*{\textbf{(a) Accuracy Changes of MTIL Datasets in Order I.}}
  \end{minipage}
  \begin{minipage}{1\textwidth}
    \centering
    \includegraphics[width=1.0\textwidth]{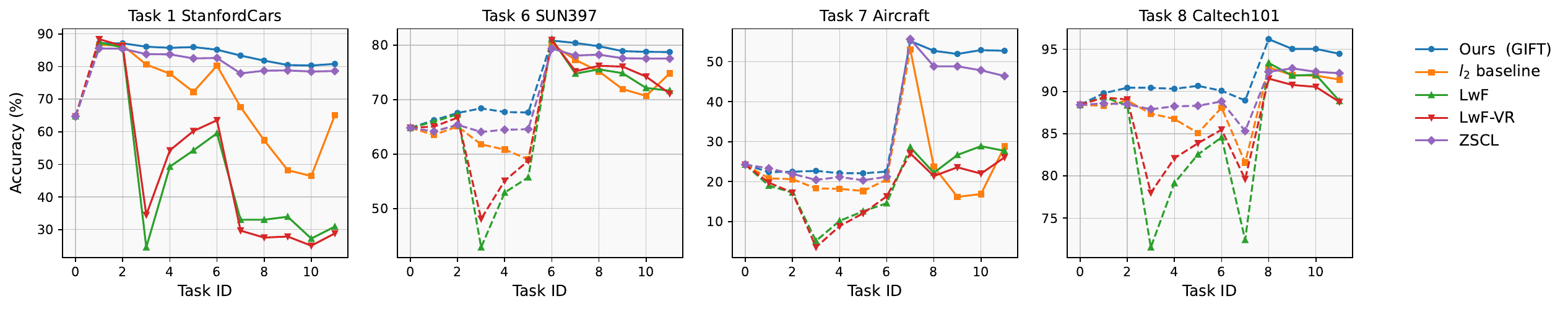}
    \vspace{-5ex}
    \caption*{\textbf{(b) Accuracy Changes of MTIL Datasets in Order II.}}
  \end{minipage}
  \caption{Illustration of the classification accuracy changes as tasks are being learned on the MTIL benchmark in two orders. The dashed lines represent the results of zero-shot predictions for an unlearned task. At task 0, the initial CLIP model's zero-shot accuracy is evaluated.}
  \vspace{-1ex}
  \label{fig:acc}
\end{figure*}

  \begin{table}[!t]
    \begin{center}
      \caption{Comparison of state-of-the-art CL methods on CIFAR100 benchmark in class-incremental setting.}
      \vspace{-4pt}
      \label{tab:cifar100_cil}
      \resizebox{\columnwidth}{!}{%
      \begin{tabular}{l|cc|cc|cc} 
      \toprule
           & \multicolumn{2}{c}{{ 10 steps}} & \multicolumn{2}{c}{{20 steps}} & \multicolumn{2}{c}{50 steps}  \\
      \textbf{Methods} & \textbf{Avg} & \textbf{Last} & \textbf{Avg} & \textbf{Last} & \textbf{Avg} & \textbf{Last} \\
      \midrule
      UCIR~\cite{hou2019learning} & 58.66 & 43.39 & 58.17 & 40.63 & 56.86 & 37.09 \\
      BiC~\cite{wu2019large} & 68.80 & 53.54 & 66.48 & 47.02 & 62.09 & 41.04 \\
      RPSNet~\cite{rajasegaran2019random}  & 68.60 & 57.05 & -  & -  & -  & -  \\
      PODNet~\cite{douillard2020podnet} & 58.03 & 41.05 & 53.97 & 35.02 & 51.19 & 32.99 \\
      DER~\cite{yoon2017lifelong} & 74.64 & 64.35 & 73.98 & 62.55 & 72.05 & 59.76 \\
      DyTox+~\cite{douillard2022dytox} & 74.10 & 62.34 & 71.62 & 57.43 & 68.90 & 51.09 \\
      \midrule
      CLIP Zero-shot& 74.47 & 65.92 & 75.20 & 65.74 & 75.67 & 65.94 \\
      CLIP Fine-tune & 65.46 & 53.23 & 59.69 & 43.13 & 39.23 & 18.89 \\
      LwF~\cite{li2017learning} & 65.86 & 48.04 & 60.64 & 40.56 & 47.69 & 32.90   \\
      iCaRL~\cite{rebuffi2017icarl} & 79.35 & 70.97 & 73.32 & 64.55 & 71.28 & 59.07 \\
      LwF-VR~\cite{ding2022don} & 78.81 & 70.75 & 74.54 & 63.54 & 71.02 & 59.45 \\
      ZSCL~\cite{zheng2023preventing} & 82.15 & 73.65 & 80.39 & 69.58  & 79.92 & 67.36  \\
      \midrule
      \rowcolor{Thistle!20}
      GIFT (Ours) & \textbf{85.11} & \textbf{77.70} & \textbf{82.11} & \textbf{73.73}  & \textbf{80.81} & \textbf{71.29}   \\
      \bottomrule
      \end{tabular}}
      \vspace{-2ex}
    \end{center}
  \end{table}

  \begin{table}[!]
    \begin{center}
      \caption{Comparison of different methods on TinyImageNet splits in class-incremental settings with 100 base classes.}
      \vspace{-4pt}
      \label{tab:tiny_imagenet_cil}
      \resizebox{\columnwidth}{!}{%
      \begin{tabular}{l|cc|cc|cc} 
    \toprule
      & \multicolumn{2}{c}{{5 steps}} &  \multicolumn{2}{c}{10 steps}    &  \multicolumn{2}{c}{20 steps}  \\
      \textbf{Methods}    & \textbf{Avg}   & \textbf{Last}    & \textbf{Avg}   & \textbf{Last}    & \textbf{Avg}   & \textbf{Last}  \\
      \midrule
      EWC~\cite{kirkpatrick2017overcoming}      & 19.01 & 6.00  & 15.82 & 3.79  & 12.35 & 4.73 \\
      EEIL~\cite{castro2018end}     & 47.17 & 35.12 & 45.03 & 34.64 & 40.41 & 29.72 \\
      UCIR~\cite{hou2019learning}     & 50.30 & 39.42 & 48.58 & 37.29 & 42.84 & 30.85 \\
      MUC~\cite{liu2020more}      & 32.23 & 19.20  & 26.67 & 15.33 & 21.89 & 10.32 \\
      PASS~\cite{zhu2021prototype}     & 49.54 & 41.64 & 47.19 & 39.27 & 42.01 & 32.93 \\
      DyTox~\cite{douillard2022dytox}    & 55.58 & 47.23 & 52.26 & 42.79 & 46.18 & 36.21 \\
      \midrule
      CLIP Zero-shot& 69.62 & 65.30 & 69.55 & 65.59 & 69.49 & 65.30 \\ 
      CLIP Fine-tune & 61.54 &46.66  &57.05 &41.54 & 54.62 & 44.55\\
      LwF~\cite{li2017learning} &60.97 &48.77 &57.60 &44.00 &54.79 &42.26 \\
      iCaRL~\cite{rebuffi2017icarl} & 77.02 & 70.39 & 73.48 & 65.97 & 69.65 & 64.68 \\
      LwF-VR~\cite{ding2022don} & 77.56 & 70.89 & 74.12 & 67.05 & 69.94 & 63.89 \\
      ZSCL~\cite{zheng2023preventing} & 80.27 & 73.57 & 78.61 & 71.62 & 77.18 & 68.30 \\
      \midrule
      \rowcolor{Thistle!20}
      GIFT (Ours) & \textbf{81.16} & \textbf{77.04} &  \textbf{80.20} & \textbf{75.51} &  \textbf{79.32}&  \textbf{74.87}\\
      \bottomrule
      \end{tabular}}
      \vspace{-3ex}
    \end{center}
  \end{table}

\noindent\textbf{Additional Benchmark Description.}
MTIL \cite{zheng2023preventing} comprises 11 datasets from diverse domains, organized into two task sequences to introduce different domain shifts. \textbf{The first sequence}, referred to as Order I, follows alphabetical order: Aircraft \cite{maji2013fine}, Caltech101 \cite{fei2004learning}, CIFAR100 \cite{krizhevsky2009learning}, DTD \cite{cimpoi2014describing}, EuroSAT \cite{helber2019eurosat}, Flowers \cite{nilsback2008automated}, Food \cite{bossard2014food}, MNIST \cite{deng2012mnist}, OxfordPet \cite{parkhi2012cats}, StanfordCars \cite{krause20133d}, SUN397 \cite{xiao2010sun}. \textbf{The second sequence}, Order II, is randomly arranged: StanfordCars, Food, MNIST, OxfordPet, Flowers, SUN397, Aircraft, Caltech101, DTD, EuroSAT, CIFAR100.

\vspace{1ex}
\noindent\textbf{Additional Metric Formulation.}
We further clarify the calculation of metrics used in the main text. Consider the accuracy matrix $[a_{i,j}]_{n \times n}$ where each element $a_{i,j}$ represents the test accuracy on task $j$ after the model has learned task $i$, evaluated across all $n$ tasks. In traditional CL, only the lower triangular portion of this matrix is relevant, as the model cannot predict for tasks it has not yet encountered. However, for VLMs, the upper triangular matrix offers valuable insight into the degradation of the model's zero-shot capability, indicating the extent of forgetting pre-training knowledge.
The metrics are computed as follows:
\begin{align}
    \text{Transfer} & = \frac{1}{n-1}\sum_{j=2}^{n}\frac{1}{j-1}\sum_{i=1}^{j-1}a_{i,j}, \\
    \text{Last} & = \frac{1}{n}\sum_{j=1}^{n}a_{n,j}, \\
    \text{Avg.} & = \frac{1}{n^2}\sum_{i=1}^{n}\sum_{j=1}^{n}a_{i,j}.
\end{align}

\vspace{1ex}
\noindent\textbf{Detailed Results.} Tab.~\ref{tab:detail_order_I} and Tab.~\ref{tab:detail_order_II} present the detailed results of Transfer, Avg, and Last metrics on each dataset of the MTIL benchmark in Order I and Order II respectively.
Task-specific metric values are calculated as described in~\cite{zheng2023preventing}, with their averages reported in the ``Average'' column, which is also presented in the main text. 
``Zero-shot'' denotes the zero-shot prediction performance of the initial CLIP model, and ``Fine-tune'' represents the direct fine-tuning accuracy on each dataset, both of which can be seen as an upper bound where no forgetting phenomenon happens. ``Continual Finetuning'' refers to the naive continual learning method that fine-tunes the model on the new task without any protection, indicating the lower bound suffering from most significant forgetting.
Evaluated under both orderings, our method can achieve the best performance on most datasets.

We select several tasks from both two orderings and plot the accuracy curves against the task ID, as shown in Fig.~\ref{fig:acc} (a) and \ref{fig:acc} (b). An ideal continual learning method for VLMs should produce an accuracy curve resembling a mirrored ``Z" shape, indicating that the zero-shot accuracy before learning the task is maintained and there is almost no forgetting after learning. Our approach aligns with this standard. However, most methods struggle with catastrophic forgetting of both pre-training knowledge and downstream task knowledge, causing significant fluctuations in the accuracy curve. This issue is particularly evident when there is a substantial domain shift, such as with the MNIST dataset (i.e., Task 8 in Fig.~\ref{fig:acc} (a) and Task 3 in Fig.~\ref{fig:acc} (b)), where the CLIP model's feature space nearly collapses without strong protection. A comparison between Fig.~\ref{fig:acc} (a) and Fig.~\ref{fig:acc} (b) shows that th the $l_2$ baseline offers some resistance to drastic domain shifts but weakens as the task sequence progresses. Since the MNIST dataset appears later in Order I than in Order II, the $l_2$ baseline can resist this domain shift in Order II but fails to do so in Order I. The strong performance of our method demonstrates that integrating knowledge distillation with the $l_2$ constraint effectively resists domain shifts and tolerates longer task sequences.

\clearpage
\begin{table*}[t!]
  \centering
  \caption{Detailed Transfer, Avg., and Last scores (\%) of different continue training methods on MTIL benchmark in \textbf{Order I}. The highest single score of each metric in each column is highlighted in \textbf{bold}, while multiple top scores are \underline{underlined}.}
  \label{tab:detail_order_I}
  {
  \begin{tabular}{@{}lcccccccccccc@{}}
  \toprule
  Method & \rotatebox{90}{Aircraft \cite{maji2013fine}} & \rotatebox{90}{Caltech101 \cite{fei2004learning}} & \rotatebox{90}{CIFAR100 \cite{krizhevsky2009learning}} & \rotatebox{90}{DTD \cite{cimpoi2014describing}} & \rotatebox{90}{EuroSAT \cite{helber2019eurosat}} & \rotatebox{90}{Flowers \cite{nilsback2008automated}} & \rotatebox{90}{Food \cite{bossard2014food}} & \rotatebox{90}{MNIST \cite{deng2012mnist}} & \rotatebox{90}{OxfordPet \cite{parkhi2012cats}} & \rotatebox{90}{Cars \cite{krause20133d}} & \rotatebox{90}{SUN397 \cite{xiao2010sun}} & Average \\ \midrule
  
   Zero-shot & 24.3 & 88.4 & 68.2 & 44.6 & 54.9 & 71.0 & 88.5 & 59.4 & 89.0 & 64.7 & 65.2 & 65.3 \\
   Fine-tune & 62.0 & 95.1 & 89.6 & 79.5 & 98.9 & 97.5 & 92.7 & 99.6 & 94.7 & 89.6 & 81.8 & 89.2 \\ \midrule
   \textbf{Transfer} \\
    Continual Finetune  & &  67.1 & 46.0 & 32.1 & 35.6 & 35.0 & 57.7 & 44.1 & 60.8 & 20.5 & 46.6 & 44.6 \\
    $l_2$ baseline & & 83.2 & 63.5 & 42.9 & 44.9 & 61.2 & 79.5 & 63.8 & 71.9 & 43.9 & 54.6 & 61.0 \\
    LwF~\cite{li2017learning} & & 74.5 & 56.9 & 39.1 & 51.1 & 52.6 & 72.8 & 60.6 & 75.1 & 30.3 & 55.9 & 56.9\\
    iCaRL~\cite{rebuffi2017icarl} & & 56.6 & 44.6 & 32.7 & 39.3 & 46.6 & 68.0 & 46.0 & 77.4 & 31.9 & 60.5 & 50.4\\
    LwF-VR~\cite{ding2022don} & & 77.1 & 61.0 & 40.5 & 45.3 & 54.4 & 74.6 & 47.9 & 76.7 & 36.3 & 58.6 & 57.2\\
    WiSE-FT~\cite{wortsman2022robust} & & 73.5 & 55.6 & 35.6 & 41.5 & 47.0 & 68.3 & 53.9 & 69.3 & 26.8 & 51.9 & 52.3\\
    ZSCL~\cite{zheng2023preventing} & & 86.0 & 67.4 & 45.4 & \textbf{50.4} & 69.1 & 87.6 & 61.8 & 86.8 & 60.1 & 66.8 & 68.1\\
    MoE-Adapter~\cite{yu2024boosting} & & 87.9 & 68.2 & 44.4 & 49.9 & \textbf{70.7} & \textbf{88.7} & 59.7 & 89.1 & \textbf{64.5} & 65.5 & 68.9\\   
    \rowcolor{Thistle!20}
    GIFT (Ours) & & \textbf{88.5} & \textbf{69.8} & \textbf{46.0} & 49.4 & 68.5 & 87.1 & \textbf{69.9} & \textbf{88.9} & 57.7 & \textbf{67.7} & \textbf{69.3}\\
    \midrule
   \textbf{Avg.} \\
    Continual Finetune & 25.5 & 81.5 & 59.1 & 53.2 & 64.7 & 51.8 & 63.2 & 64.3 & 69.7 & 31.8 & 49.7 & 55.9 \\
    $l_2$ baseline & 24.0 & 82.3 & 68.2 & 58.2 & 70.9 & 67.0 & 76.2 & 57.1 & 77.5 & 51.4 & 56.9 & 62.7 \\
    LwF~\cite{li2017learning} & 36.3 & 86.9 & 72.0 & 59.0 & 73.7 & 60.0 & 73.6 & 74.8 & 80.0 & 37.3 & 58.1 & 64.7 \\
    iCaRL~\cite{rebuffi2017icarl} & 35.5 & 89.2 & 72.2 & 60.6 & 68.8 & 70.0 & 78.2 & 62.3 & 81.8 & 41.2 & 62.5 & 65.7 \\
    LwF-VR~\cite{ding2022don} & 29.6 & 87.7 & 74.4 & 59.5 & 72.4 & 63.6 & 77.0 & 66.7 & 81.2 & 43.7 & 60.7 & 65.1 \\
    WiSE-FT~\cite{wortsman2022robust} & 26.7 & 86.5 & 64.3 & 57.1 & 65.7 & 58.7 & 71.1 & 70.5 & 75.8 & 36.9 & 54.6 & 60.7\\
    ZSCL~\cite{zheng2023preventing} & 45.1 & 92.0 & 80.1 & 64.3 & 79.5 & 81.6 & \textbf{89.6} & 75.2 & 88.9 & 64.7 & 68.0 &75.4 \\
    MoE-Adapter~\cite{yu2024boosting} & 50.2 & 91.9 & \textbf{83.1} & \textbf{69.4} & 78.9 & \textbf{84.0} & 89.1 & 73.7 & 89.3 & \textbf{67.7} & 66.9 & 76.7 \\ 
    \rowcolor{Thistle!20}
    GIFT (Ours) & \textbf{51.9} & \textbf{93.9} & 81.4 & 67.7 & \textbf{80.3} & 82.8 & 89.3 & \textbf{80.6} & \textbf{90.3} & 63.1 & \textbf{68.9} & \textbf{77.3}\\
    \midrule
   \textbf{Last} \\
    Continual Finetune & 31.0 & 89.3 & 65.8 & 67.3 & 88.9 & 71.1 & 85.6 & \underline{99.6} & 92.9 & 77.3 & 81.1 & 77.3\\
    $l_2$ baseline & 22.4 & 91.1 & 80.8 & 69.2 & 93.5 & 81.2 & 90.5 & 49.4 & 92.7 & 83.8 & 80.1 & 75.9 \\
    LwF~\cite{li2017learning} & 26.3 & 87.5 & 71.9 & 66.6 & 79.9 & 66.9 & 83.8 & \underline{99.6} & 92.1 & 66.1 & 80.4 & 74.6 \\
    iCaRL~\cite{rebuffi2017icarl} & 35.8 & 93.0 & 77.0 & 70.2 & 83.3 & 88.5 & 90.4 & 86.7 & 93.2 & 81.2 & \textbf{81.9} & 80.1\\
    LwF-VR~\cite{ding2022don} & 20.5 & 89.8 & 72.3 & 67.6 & 85.5 & 73.8 & 85.7 & \underline{99.6} & 93.1 & 73.3 & 80.9 & 76.6 \\
    WiSE-FT~\cite{wortsman2022robust} & 27.2 & 90.8 & 68.0 & 68.9 & 86.9 & 74.0 & 87.6 & \underline{99.6} & 92.6 & 77.8 & 81.3 & 77.7\\
    ZSCL~\cite{zheng2023preventing} & 40.6 & 92.2 & 81.3 & 70.5 & 94.8 & 90.5 & \textbf{91.9} & 98.7 & 93.9 & 85.3 & 80.2 & 83.6\\
    MoE-Adapter~\cite{yu2024boosting} & \textbf{49.8} & 92.2 & \textbf{86.1} & \textbf{78.1} & 95.7 & \textbf{94.3} & 89.5 & 98.1 & 89.9 & 81.6 & 80.0 & 85.0 \\
    \rowcolor{Thistle!20}
    GIFT (Ours) & 47.9 & \textbf{95.6} & 82.8 & 75.1 & \textbf{97.3} & 94.2 & 91.7 & 99.2 & \textbf{94.2} & \textbf{87.0} & 80.9 & \textbf{86.0}\\
  \bottomrule
  \end{tabular}%
  }
  \end{table*}
\clearpage
\begin{table*}[t!]
  \centering
  \caption{Detailed Transfer, Avg., Last accuracy (\%) of different continue training methods on MTIL benchmark in \textbf{Order II}. The highest single score of each metric in each column is highlighted in \textbf{bold}, while multiple top scores are \underline{underlined}.}
  \label{tab:detail_order_II}
  {
  \begin{tabular}{@{}lcccccccccccc@{}}
  \toprule
  Method & \rotatebox{90}{Cars \cite{krause20133d}} & \rotatebox{90}{Food \cite{bossard2014food}} & \rotatebox{90}{MNIST \cite{deng2012mnist}} & \rotatebox{90}{OxfordPet \cite{parkhi2012cats}} & \rotatebox{90}{Flowers \cite{nilsback2008automated}} & \rotatebox{90}{SUN397 \cite{xiao2010sun}} & \rotatebox{90}{Aircraft \cite{maji2013fine}} & \rotatebox{90}{Caltech101 \cite{fei2004learning}} & \rotatebox{90}{DTD \cite{cimpoi2014describing}} & \rotatebox{90}{EuroSAT \cite{helber2019eurosat}} & \rotatebox{90}{CIFAR100 \cite{krizhevsky2009learning}} & Average\\ \midrule
  
   Zero-shot & 64.7 & 88.5 & 59.4 & 89.0 & 71.0 & 65.2 & 24.3 & 88.4 & 44.6 & 54.9 & 68.2 & 65.3 \\
   Fine-tune & 89.6 & 92.7 & 94.7 & 94.7 & 97.5 & 81.8 & 62.0 & 95.1 & 79.5 & 98.9 & 89.6 & 89.2 \\ \midrule
   \textbf{Transfer} \\
    Continual Finetune & & 85.9 & 59.6 & 57.9 & 40.0 & 46.7 & 11.1 & 70.0 & 30.5 & 26.6 & 37.7 & 46.6 \\
    $l_2$ baseline & & 87.0 & 62.3 & 83.7 & 60.6 & 62.1 & 19.3 & 86.6 & 42.7 & 41.4 & 60.1 & 60.6 \\
    LwF~\cite{li2017learning}  & & 87.8 & 58.5 & 71.9 & 46.6 & 57.3 & 12.8 & 81.4 & 34.5 & 34.5 & 46.8 & 53.2 \\
    iCaRL~\cite{rebuffi2017icarl} & & 86.1 & 51.8 & 67.6 & 50.4 & 57.9 & 11.0 & 72.3 & 31.2 & 32.7 & 48.1 & 50.9 \\
    LwF-VR~\cite{ding2022don}  & & 88.2 & 57.0 & 71.4 & 50.0 & 58.0 & 13.0 & 82.0 & 34.4 & 29.3 & 47.6 & 53.1 \\
    WiSE-FT~\cite{wortsman2022robust}  & & 87.2 & 57.6 & 67.0 & 45.0 & 54.0 & 12.9 & 78.6 & 35.5 & 28.4 & 44.3 & 51.0 \\
    ZSCL~\cite{zheng2023preventing}  & & 88.3 & 57.5 & 84.7 & 68.1 & 64.8 & 21.1 & 88.2 & 45.3 & \textbf{55.2} & 68.2 & 64.2 \\
    MoE-Adapter~\cite{yu2024boosting} & & \textbf{88.8} & 59.5 & \textbf{89.1} & 69.9 & 64.4 & 18.1 & 86.9 & 43.7 & 54.6 & 68.2 & 64.3 \\
    \rowcolor{Thistle!20}
    GIFT (Ours) & & 88.3 & \textbf{63.4} & 88.1 & \textbf{70.8} & \textbf{67.7} & \textbf{22.8} & \textbf{90.4} & \textbf{46.7} & 51.8 & \textbf{68.8} & \textbf{65.9} \\
    \midrule
  
   \textbf{Avg.} \\
    Continual Finetune  & 42.1 & 70.5 & 92.2 & 80.1 & 54.5 & 59.1 & 19.8 & 78.3 & 41.0 & 38.1 & 42.3 & 56.2 \\
    $l_2$ baseline & 69.9 & 86.2 & 91.9 & 89.0 & 74.0 & 69.1 & 23.2 & 88.6 & 51.1 & 50.8 & 62.6 & 68.8 \\
    LwF~\cite{li2017learning}  & 49.0 & 77.0 & 92.1 & 85.9 & 66.5 & 67.2 & 20.9 & 84.7 & 44.6 & 45.5 & 50.5 & 62.2 \\
    iCaRL~\cite{rebuffi2017icarl} & 52.0 & 75.9 & 77.4 & 74.6 & 58.4 & 59.3 & 11.7 & 79.6 & 42.1 & 43.2 & 51.7 & 56.9 \\
    LwF-VR~\cite{ding2022don}  & 44.9 & 75.8 & 91.8 & 85.3 & 63.5 & 67.6 & 16.9 & 84.9 & 44.0 & 40.6 & 51.3 & 60.6 \\
    WiSE-FT~\cite{wortsman2022robust}  & 52.6 & 79.3 & 91.9 & 83.9 & 63.4 & 65.2 & 23.3 & 83.7 & 45.4 & 40.0 & 48.2 & 61.5 \\
    ZSCL~\cite{zheng2023preventing} & 81.7 & \textbf{91.3} & 91.1 & 91.0 & 82.9 & 72.5 & 33.6 & 89.7 & 53.3 & \textbf{62.8} & 69.9 & 74.5\\
    MoE-Adapter~\cite{yu2024boosting} & \textbf{84.9} & 89.9 & 89.3 & 91.4 & \textbf{86.2} & 72.2 & 33.4 & 89.4 & 53.3 & 61.4 & 69.9 & 74.7 \\
    \rowcolor{Thistle!20}
    GIFT (Ours) & 83.2 & 90.8 & \textbf{92.6} & \textbf{92.8} & 85.8 & \textbf{74.1} & \textbf{36.0} & \textbf{92.1} & \textbf{54.7} & 60.0 & \textbf{70.4} & \textbf{75.7} \\
    \midrule
  
   \textbf{Last} \\
    Continual Finetune  & 24.0 & 67.3 & 99.1 & 87.4 & 44.3 & 67.0 & 29.5 & 92.3 & 61.3 & 81.0 & 88.1 & 67.4 \\
    $l_2$ baseline & 65.1 & 84.2 & 96.4 & 90.2 & 71.8 & 74.8 & 28.8 & 91.4 & 70.7 & 88.2 & 87.2 & 77.2 \\
    LwF~\cite{li2017learning} & 34.6 & 69.6 & 99.3 & 88.7 & 61.1 & 72.5 & 32.5 & 88.1 & 65.6 & 90.9 & 87.9 & 71.9 \\
    iCaRL~\cite{rebuffi2017icarl} & 46.0 & 81.5 & 91.3 & 82.8 & 66.5 & 72.2 & 16.3 & 91.6 & 68.1 & 83.2 & 87.8 & 71.6 \\
    LwF-VR~\cite{ding2022don} & 27.4 & 61.2 & 99.4 & 86.3 & 60.6 & 70.7 & 23.4 & 88.0 & 61.3 & 84.3 & 88.1 & 68.3 \\
    WiSE-FT~\cite{wortsman2022robust} & 35.6 & 76.9 & \textbf{99.5} & 89.1 & 62.1 & 71.8 & 27.8 & 90.8 & 67.0 & 85.6 & \textbf{87.6} & 72.2 \\
    ZSCL~\cite{zheng2023preventing} & 78.2 & \textbf{91.1} & 97.6 & 92.5 & 87.4 & 78.2 & 45.0 & 92.3 & 72.7 & \textbf{96.2} & 86.3 &83.4\\
    MoE-Adapter~\cite{yu2024boosting} & \textbf{84.1} & 88.5 & 94.0 & 91.8 & \textbf{94.1} & 77.8 & 50.4 & 93.3 & \textbf{77.1} & 87.7 & 86.6 & 84.1 \\
    \rowcolor{Thistle!20}
    GIFT (Ours) & 81.0 & 90.2 & 98.6 & \textbf{94.0} & 91.5 & \textbf{78.6} & \textbf{51.7} & \textbf{94.6} & 75.6 & 95.4 & 86.6 & \textbf{85.3} \\
  \bottomrule
  \end{tabular}%
  }
  \end{table*}
\clearpage

\section*{D. Ablation Analysis of Image Generation}
In this section, we conduct an ablation study on the image generation mechanism, primarily on the hyperparameters of Stable Diffusion inference, i.e., denoising steps and classifier-free guidance scale ~\cite{ho2022classifier}. Additionally, we examine the impact of eliminating synthetic images of specific downstream task datasets on the distillation performance.

\vspace{1ex}
\noindent\textbf{Denoising Steps.}
For Stable Diffusion, the number of denoising steps is a crucial hyperparameter that balances generation speed and quality. Fewer denoising steps result in faster generation but typically at the cost of lower quality. While our method defaults to 50 denoising steps, it remains effective with lower settings, such as 25 denoising steps, for faster generation. As shown in Tab.~\ref{tab:denoising_steps}, reducing the steps to 25 has minimal impact on performance of our method. By fixing the random seed, the images generated with fewer steps correspond to intermediate outputs from more denoising steps. We further visualize and compare the quality of images generated with different denoising steps in our synthetic dataset, as illustrated in Fig.~\ref{fig:image_source}.

\begin{table}[h!]
  \centering
  \caption{Comparison of synthetic images generated with different denosing steps as data sources for distillation.}
  \resizebox{0.95\columnwidth}{!}{%
  \begin{tabular}{@{}l|c|c|c|c@{}}
      \toprule
      Method & Denoising Steps & Transfer & Avg. & Last  \\
      \midrule
      GIFT w/ AWC & 50 Steps  & \textbf{69.3} & \textbf{77.3} & \textbf{86.0}\\
      GIFT w/o AWC & 50 Steps  & 68.9 & 76.6 & 85.0\\
      \midrule
      GIFT w/ AWC & 25 Steps  &69.2 &77.2 &85.8 \\
      GIFT w/o AWC & 25 Steps  & 69.2 &76.6 &84.8 \\
      \bottomrule
  \end{tabular}%
  }
  \label{tab:denoising_steps}
  \vspace{-2ex}
\end{table}

\vspace{1ex}
\noindent\textbf{Classifier-free Guidance Scale.} We consider three configurations for the classifier-free guidance scale $w$: large scale ($w=10.5$), medium scale ($w=7.5$) and small scale ($w=4.5$). Prior studies~\cite{tian2024stablerep} suggest that when training with large-scale synthetic images, smaller guidance scales are crucial for enhancing the diversity of generated images and boosting performance, because smaller $w$ leads to greater intra-caption variation between generated images. However, as shown in Tab.~\ref{tab:guidance_scale}, our method demonstrates consistent performance across different guidance scales, with slightly reduced effectiveness observed for smaller scales. This could be attributed to the fact that when fewer synthetic images are used, \textbf{inter-class diversity is more critical than intra-class diversity}. We ensure inter-class diversity by constructing distinct prompts based on different class names. Meanwhile, the increased deviation between the generated images and the text prompts with a small $w$ may negatively affect the memory retention of CLIP, as evidenced by the lower Transfer value.

\begin{figure*}[h!]
  \centering
  \begin{minipage}{1.0\textwidth}
    \centering
    \includegraphics[width=\textwidth]{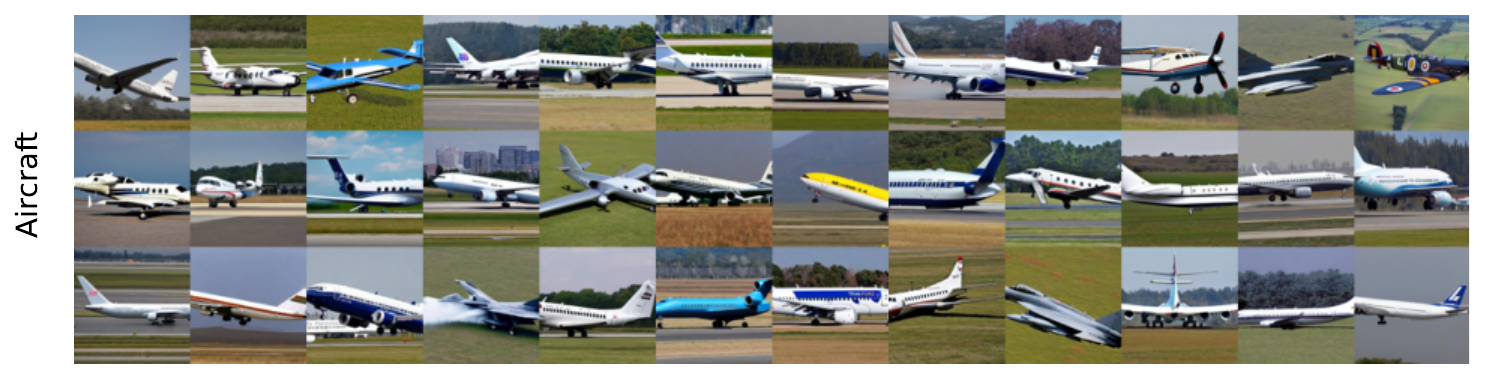}
    \vspace{-15pt}
  \end{minipage}
  \begin{minipage}{1.0\textwidth}
    \centering
    \includegraphics[width=\textwidth]{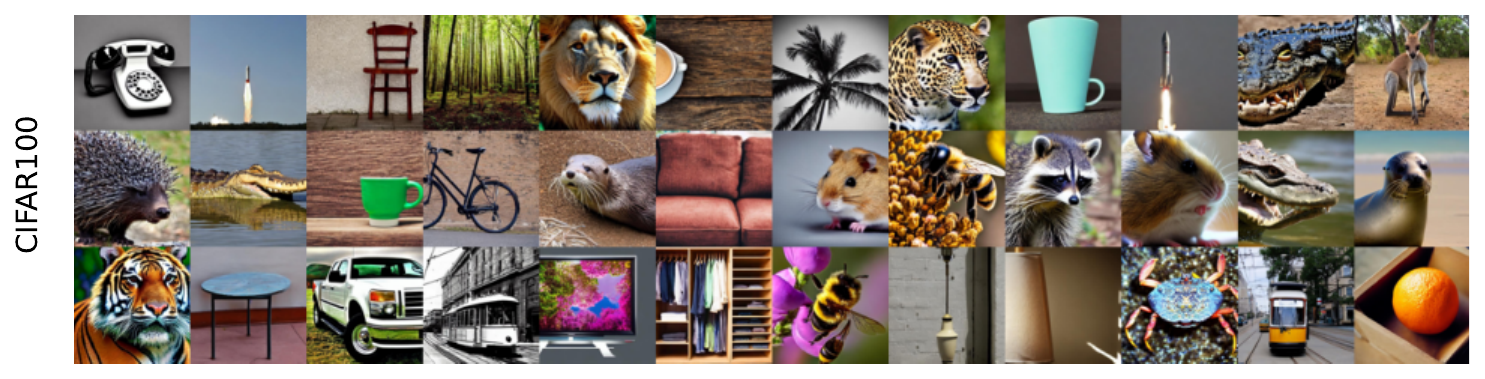}
    \vspace{-15pt}
  \end{minipage}
  \begin{minipage}{1.0\textwidth}
    \centering
    \includegraphics[width=\textwidth]{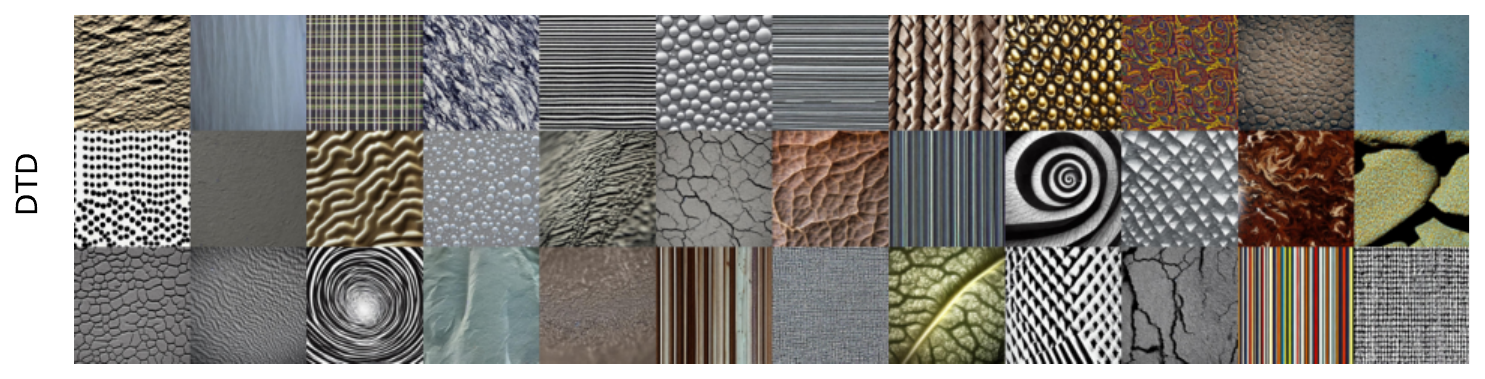}
    \vspace{-15pt}
  \end{minipage}
  \begin{minipage}{1.0\textwidth}
    \centering
    \includegraphics[width=\textwidth]{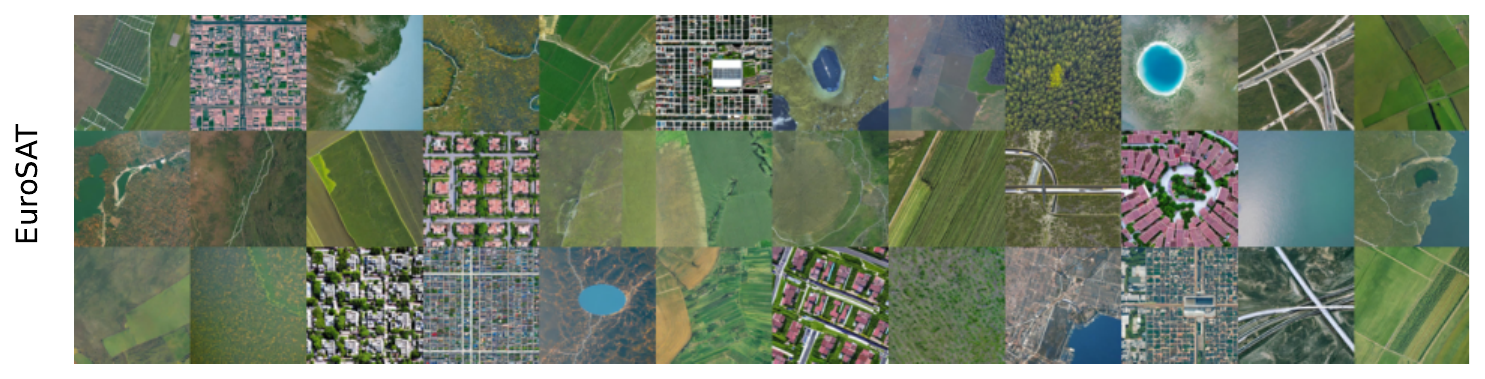}
    \vspace{-15pt}
  \end{minipage}
  \begin{minipage}{1.0\textwidth}
    \centering
    \includegraphics[width=\textwidth]{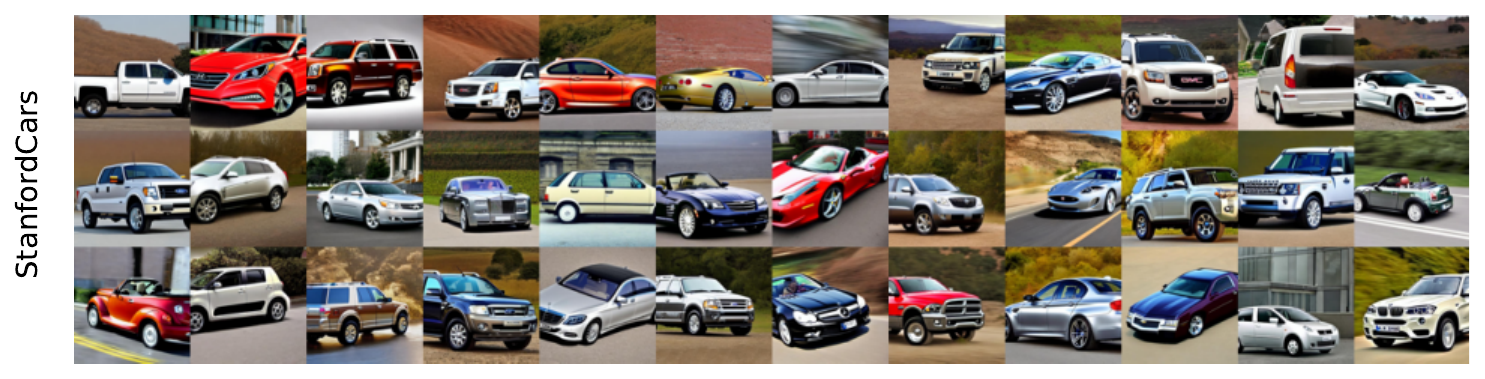}
    \vspace{-15pt}
  \end{minipage}
  \caption{Visualization of synthetic data for different downstream datasets.}
  \label{fig:syn_sample}
\end{figure*}

\vspace{1ex}
\noindent\textbf{Eliminating Synthetic Images for Downstream Tasks.} We skip the class names of specific downstream dataset so that the classes included in this dataset do not appear in the synthetic images. In doing so, we conduct experiments on skipping the generation of the first datasets of MTIL in order I and order II, i.e., the Aircraft and StanfordCars datasets. These two datasets are selected because they appear first in the sequence and are fine-grained, making them more susceptible to forgetting during continual learning. As can be seen from the accuracy curves in Fig. ~\ref{fig:syn_sample}, incorporating synthetic data for a specific downstream task significantly enhances memory retention of that task.

\begin{table}[t]
  \centering
  \caption{Comparison of synthetic images generated with different classifier-free guidance scale as data sources for distillation.}
  \resizebox{0.88\columnwidth}{!}{%
  \begin{tabular}{@{}c|c|c|c|c@{}}
      \toprule
      Guidance Scale & Image Num & Transfer & Avg. & Last  \\
      \midrule
      small & \multirow{3}{*}{1K} & 68.2 & 76.3 & 85.2 \\
      medium &                   & 68.9 & 76.6 & 85.0 \\
      large &                    & 68.5 & 76.3 & 85.1 \\
      \midrule
      small & \multirow{3}{*}{3K} & 68.7 & 76.8 & 85.0 \\
      medium &                   & 69.1 & 76.7 & 84.9 \\
      large &                    & 68.8 & 76.6 & 85.1 \\
      \bottomrule
  \end{tabular}%
  }
  \label{tab:guidance_scale}
\end{table}

\begin{figure}[t]
  \centering
  \includegraphics[width=1\columnwidth]{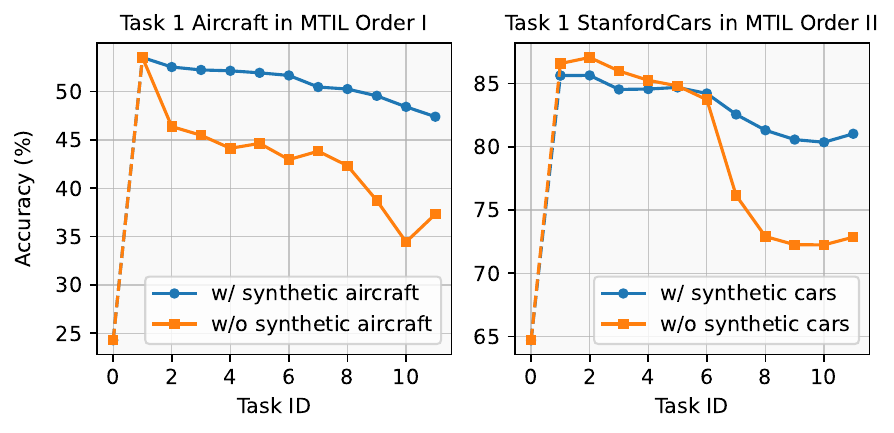}
  \caption{Eliminating synthetic images for specific downstream tasks exacerbates forgetting of these tasks.}
  \vspace{-3ex}
  \label{fig:downstream}
\end{figure}

\section*{E. Visualization: Synthetic Data for Different Datasets}
\begin{figure*}[h!]
  \centering
  \begin{minipage}{1.0\textwidth}
    \centering
    \includegraphics[width=\textwidth]{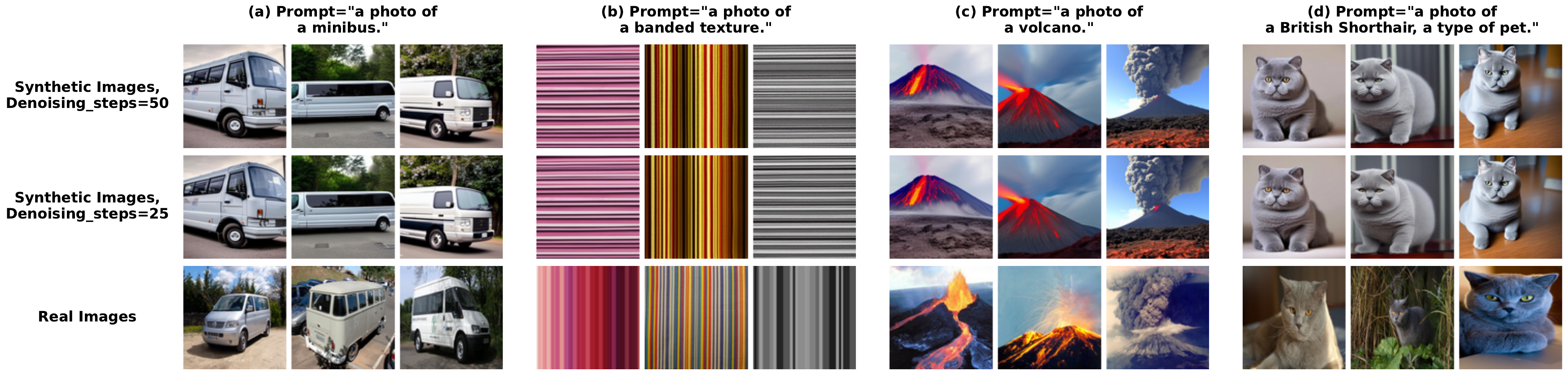}
    \vspace{-5pt}
  \end{minipage}
  \begin{minipage}{1.0\textwidth}
    \centering
    \includegraphics[width=\textwidth]{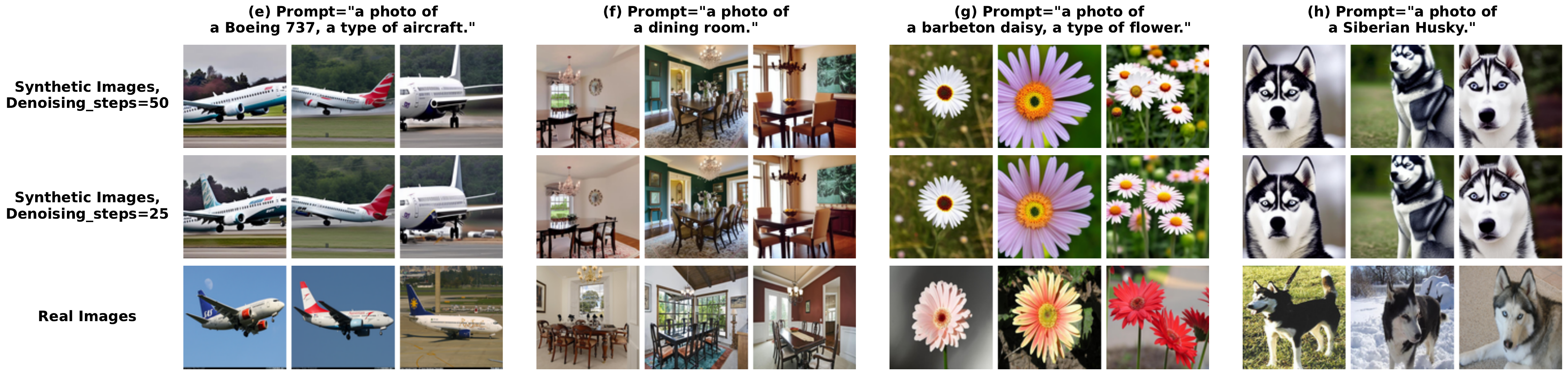}
    \vspace{-5pt}
  \end{minipage}
  \begin{minipage}{1.0\textwidth}
    \centering
    \includegraphics[width=\textwidth]{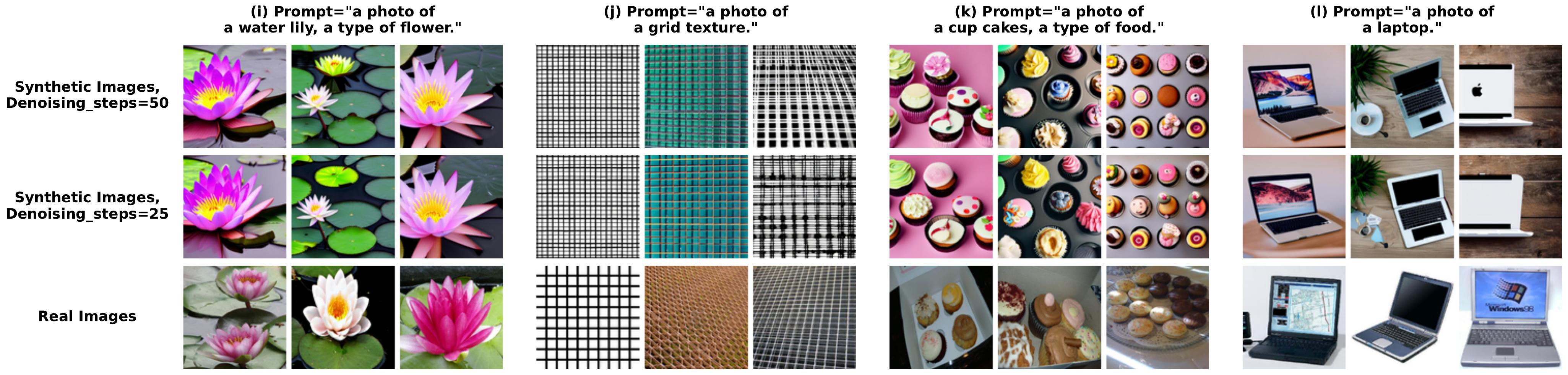}
    \vspace{-5pt}
  \end{minipage}
  \begin{minipage}{1.0\textwidth}
    \centering
    \includegraphics[width=\textwidth]{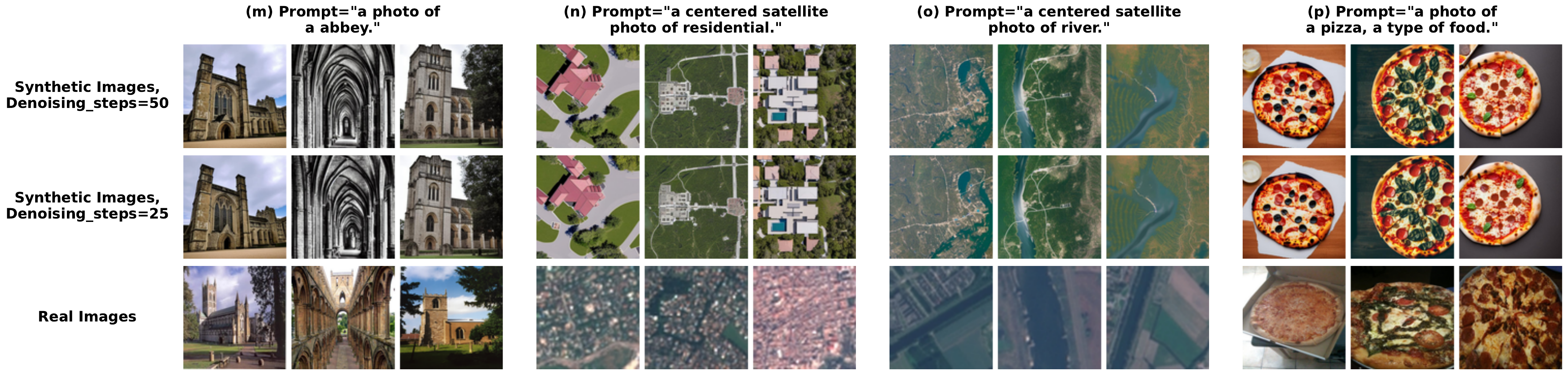}
    \vspace{-13pt}
  \end{minipage}
  \caption{Leveraging the high-quality image generation capabilities of Stable Diffusion, we generate diverse and vivid images across different categories using simple textual prompts. Reducing the number of denoising steps to 25 doesn't bring much visual degradation of the generated images. Our comparative experiments demonstrate that our method remains effective with fewer denoising steps, allowing for faster generation for practical applications.}
  \vspace{-10pt}
  \label{fig:image_source}
\end{figure*}

At last, we provide synthetic images for different downstream datasets (i.e., Aircraft, CIFAR100, DTD, EuroSAT and StanfordCars) in Fig.~\ref{fig:image_source}. All images are randomly chosen rather than human-picked and are used in our experiments. We observe that for most datasets, synthesized images from the Stable Diffusion model are of high quality, demonstrating its capability to adapt to diverse domains without fine-tuning. But there also exist cases that many unsatisfactory examples are generated, such as the DTD and EuroSAT datasets. Additionally, Stable Diffusion struggles with generating accurate numerical representations, making it unsuitable for datasets like MNIST. It also fails to replicate the low-resolution nature of CIFAR-100 images but successfully captures the classes within the dataset.  Despite these limitations, the majority of the generated images are satisfactory, underscoring the significant contribution of Stable Diffusion to overcoming catastrophic forgetting.

\clearpage

\clearpage

%% file: main.bbl
\begin{thebibliography}{82}
\providecommand{\natexlab}[1]{#1}
\providecommand{\url}[1]{\texttt{#1}}
\expandafter\ifx\csname urlstyle\endcsname\relax
  \providecommand{\doi}[1]{doi: #1}\else
  \providecommand{\doi}{doi: \begingroup \urlstyle{rm}\Url}\fi

\bibitem[Aljundi et~al.(2017)Aljundi, Chakravarty, and Tuytelaars]{aljundi2017expert}
Rahaf Aljundi, Punarjay Chakravarty, and Tinne Tuytelaars.
\newblock Expert gate: Lifelong learning with a network of experts.
\newblock In \emph{CVPR}, 2017.

\bibitem[Aljundi et~al.(2018)Aljundi, Babiloni, Elhoseiny, Rohrbach, and Tuytelaars]{aljundi2018memory}
Rahaf Aljundi, Francesca Babiloni, Mohamed Elhoseiny, Marcus Rohrbach, and Tinne Tuytelaars.
\newblock Memory aware synapses: Learning what (not) to forget.
\newblock In \emph{ECCV}, 2018.

\bibitem[Aljundi et~al.(2019)Aljundi, Lin, Goujaud, and Bengio]{aljundi2019gradient}
Rahaf Aljundi, Min Lin, Baptiste Goujaud, and Yoshua Bengio.
\newblock Gradient based sample selection for online continual learning.
\newblock \emph{NeurIPS}, 2019.

\bibitem[Bang et~al.(2021)Bang, Kim, Yoo, Ha, and Choi]{bang2021rainbow}
Jihwan Bang, Heesu Kim, YoungJoon Yoo, Jung-Woo Ha, and Jonghyun Choi.
\newblock Rainbow memory: Continual learning with a memory of diverse samples.
\newblock In \emph{CVPR}, 2021.

\bibitem[Bossard et~al.(2014)Bossard, Guillaumin, and Van~Gool]{bossard2014food}
Lukas Bossard, Matthieu Guillaumin, and Luc Van~Gool.
\newblock Food-101--mining discriminative components with random forests.
\newblock In \emph{ECCV}, 2014.

\bibitem[Buzzega et~al.(2020)Buzzega, Boschini, Porrello, Abati, and Calderara]{buzzega2020dark}
Pietro Buzzega, Matteo Boschini, Angelo Porrello, Davide Abati, and Simone Calderara.
\newblock Dark experience for general continual learning: a strong, simple baseline.
\newblock \emph{NeurIPS}, 2020.

\bibitem[Castro et~al.(2018)Castro, Mar{\'\i}n-Jim{\'e}nez, Guil, Schmid, and Alahari]{castro2018end}
Francisco~M Castro, Manuel~J Mar{\'\i}n-Jim{\'e}nez, Nicol{\'a}s Guil, Cordelia Schmid, and Karteek Alahari.
\newblock End-to-end incremental learning.
\newblock In \emph{ECCV}, 2018.

\bibitem[Cimpoi et~al.(2014)Cimpoi, Maji, Kokkinos, Mohamed, and Vedaldi]{cimpoi2014describing}
Mircea Cimpoi, Subhransu Maji, Iasonas Kokkinos, Sammy Mohamed, and Andrea Vedaldi.
\newblock Describing textures in the wild.
\newblock In \emph{CVPR}, 2014.

\bibitem[Csisz{\'a}r(1975)]{csiszar1975divergence}
Imre Csisz{\'a}r.
\newblock I-divergence geometry of probability distributions and minimization problems.
\newblock \emph{The Annals of Probability}, 1975.

\bibitem[Deng et~al.(2009)Deng, Dong, Socher, Li, Li, and Fei-Fei]{deng2009imagenet}
Jia Deng, Wei Dong, Richard Socher, Li-Jia Li, Kai Li, and Li Fei-Fei.
\newblock Imagenet: A large-scale hierarchical image database.
\newblock In \emph{CVPR}, 2009.

\bibitem[Deng(2012)]{deng2012mnist}
Li Deng.
\newblock The mnist database of handwritten digit images for machine learning research.
\newblock \emph{IEEE Signal Processing Magazine}, 2012.

\bibitem[Ding et~al.(2022)Ding, Liu, Tian, Yang, and Ding]{ding2022don}
Yuxuan Ding, Lingqiao Liu, Chunna Tian, Jingyuan Yang, and Haoxuan Ding.
\newblock Don't stop learning: Towards continual learning for the clip modeltoward.
\newblock \emph{arXiv preprint arXiv:2207.09248}, 2022.

\bibitem[Dosovitskiy et~al.(2020)Dosovitskiy, Beyer, Kolesnikov, Weissenborn, Zhai, Unterthiner, Dehghani, Minderer, Heigold, Gelly, et~al.]{dosovitskiy2020image}
Alexey Dosovitskiy, Lucas Beyer, Alexander Kolesnikov, Dirk Weissenborn, Xiaohua Zhai, Thomas Unterthiner, Mostafa Dehghani, Matthias Minderer, Georg Heigold, Sylvain Gelly, et~al.
\newblock An image is worth 16x16 words: Transformers for image recognition at scale.
\newblock \emph{arXiv preprint arXiv:2010.11929}, 2020.

\bibitem[Douillard et~al.(2020)Douillard, Cord, Ollion, Robert, and Valle]{douillard2020podnet}
Arthur Douillard, Matthieu Cord, Charles Ollion, Thomas Robert, and Eduardo Valle.
\newblock Podnet: Pooled outputs distillation for small-tasks incremental learning.
\newblock In \emph{ECCV}, 2020.

\bibitem[Douillard et~al.(2022)Douillard, Ram{\'e}, Couairon, and Cord]{douillard2022dytox}
Arthur Douillard, Alexandre Ram{\'e}, Guillaume Couairon, and Matthieu Cord.
\newblock Dytox: Transformers for continual learning with dynamic token expansion.
\newblock In \emph{CVPR}, 2022.

\bibitem[Fei-Fei et~al.(2004)Fei-Fei, Fergus, and Perona]{fei2004learning}
Li Fei-Fei, Rob Fergus, and Pietro Perona.
\newblock Learning generative visual models from few training examples: An incremental bayesian approach tested on 101 object categories.
\newblock In \emph{CVPRW}, 2004.

\bibitem[Gao et~al.(2024)Gao, Geng, Zhang, Ma, Fang, Zhang, Li, and Qiao]{gao2024clip}
Peng Gao, Shijie Geng, Renrui Zhang, Teli Ma, Rongyao Fang, Yongfeng Zhang, Hongsheng Li, and Yu Qiao.
\newblock Clip-adapter: Better vision-language models with feature adapters.
\newblock \emph{IJCV}, 2024.

\bibitem[Gao and Liu(2023)]{gao2023ddgr}
Rui Gao and Weiwei Liu.
\newblock Ddgr: Continual learning with deep diffusion-based generative replay.
\newblock In \emph{ICML}, 2023.

\bibitem[Garipov et~al.(2018)Garipov, Izmailov, Podoprikhin, Vetrov, and Wilson]{garipov2018loss}
Timur Garipov, Pavel Izmailov, Dmitrii Podoprikhin, Dmitry~P Vetrov, and Andrew~G Wilson.
\newblock Loss surfaces, mode connectivity, and fast ensembling of dnns.
\newblock \emph{NeurIPS}, 2018.

\bibitem[Goyal et~al.(2023)Goyal, Kumar, Garg, Kolter, and Raghunathan]{goyal2023finetune}
Sachin Goyal, Ananya Kumar, Sankalp Garg, Zico Kolter, and Aditi Raghunathan.
\newblock Finetune like you pretrain: Improved finetuning of zero-shot vision models.
\newblock In \emph{CVPR}, 2023.

\bibitem[He et~al.(2018)He, Wang, Shan, and Chen]{he2018exemplar}
Chen He, Ruiping Wang, Shiguang Shan, and Xilin Chen.
\newblock Exemplar-supported generative reproduction for class incremental learning.
\newblock In \emph{BMVC}, 2018.

\bibitem[He et~al.(2016)He, Zhang, Ren, and Sun]{he2016deep}
Kaiming He, Xiangyu Zhang, Shaoqing Ren, and Jian Sun.
\newblock Deep residual learning for image recognition.
\newblock In \emph{CVPR}, 2016.

\bibitem[He et~al.(2023)He, Sun, Yu, Xue, Zhang, Torr, Bai, and QI]{he2023synthetic}
Ruifei He, Shuyang Sun, Xin Yu, Chuhui Xue, Wenqing Zhang, Philip Torr, Song Bai, and XIAOJUAN QI.
\newblock Is synthetic data from generative models ready for image recognition?
\newblock In \emph{ICLR}, 2023.

\bibitem[Helber et~al.(2019)Helber, Bischke, Dengel, and Borth]{helber2019eurosat}
Patrick Helber, Benjamin Bischke, Andreas Dengel, and Damian Borth.
\newblock Eurosat: A novel dataset and deep learning benchmark for land use and land cover classification.
\newblock \emph{IEEE Journal of Selected Topics in Applied Earth Observations and Remote Sensing}, 2019.

\bibitem[Hessel et~al.(2021)Hessel, Holtzman, Forbes, Bras, and Choi]{hessel2021clipscore}
Jack Hessel, Ari Holtzman, Maxwell Forbes, Ronan~Le Bras, and Yejin Choi.
\newblock Clipscore: A reference-free evaluation metric for image captioning.
\newblock \emph{arXiv preprint arXiv:2104.08718}, 2021.

\bibitem[Hinton et~al.(2015)Hinton, Vinyals, and Dean]{hinton2015distilling}
Geoffrey Hinton, Oriol Vinyals, and Jeff Dean.
\newblock Distilling the knowledge in a neural network.
\newblock \emph{arXiv preprint arXiv:1503.02531}, 2015.

\bibitem[Ho and Salimans(2022)]{ho2022classifier}
Jonathan Ho and Tim Salimans.
\newblock Classifier-free diffusion guidance.
\newblock \emph{arXiv preprint arXiv:2207.12598}, 2022.

\bibitem[Ho et~al.(2020)Ho, Jain, and Abbeel]{ho2020denoising}
Jonathan Ho, Ajay Jain, and Pieter Abbeel.
\newblock Denoising diffusion probabilistic models.
\newblock \emph{NeurIPS}, 2020.

\bibitem[Hou et~al.(2019)Hou, Pan, Loy, Wang, and Lin]{hou2019learning}
Saihui Hou, Xinyu Pan, Chen~Change Loy, Zilei Wang, and Dahua Lin.
\newblock Learning a unified classifier incrementally via rebalancing.
\newblock In \emph{CVPR}, 2019.

\bibitem[Huang et~al.(2024)Huang, Liang, Shi, Zhu, Wan, Li, Du, Tao, and Ye]{huang2024learn}
Wenke Huang, Jian Liang, Zekun Shi, Didi Zhu, Guancheng Wan, He Li, Bo Du, Dacheng Tao, and Mang Ye.
\newblock Learn from downstream and be yourself in multimodal large language model fine-tuning.
\newblock \emph{arXiv preprint arXiv:2411.10928}, 2024.

\bibitem[Jia et~al.(2021)Jia, Yang, Xia, Chen, Parekh, Pham, Le, Sung, Li, and Duerig]{jia2021scaling}
Chao Jia, Yinfei Yang, Ye Xia, Yi-Ting Chen, Zarana Parekh, Hieu Pham, Quoc Le, Yun-Hsuan Sung, Zhen Li, and Tom Duerig.
\newblock Scaling up visual and vision-language representation learning with noisy text supervision.
\newblock In \emph{ICML}, 2021.

\bibitem[Jia et~al.(2022)Jia, Tang, Chen, Cardie, Belongie, Hariharan, and Lim]{jia2022visual}
Menglin Jia, Luming Tang, Bor-Chun Chen, Claire Cardie, Serge Belongie, Bharath Hariharan, and Ser-Nam Lim.
\newblock Visual prompt tuning.
\newblock In \emph{ECCV}, 2022.

\bibitem[Jodelet et~al.(2023)Jodelet, Liu, Phua, and Murata]{jodelet2023class}
Quentin Jodelet, Xin Liu, Yin~Jun Phua, and Tsuyoshi Murata.
\newblock Class-incremental learning using diffusion model for distillation and replay.
\newblock In \emph{ICCVW}, 2023.

\bibitem[Kim et~al.(2024)Kim, Cho, Kim, Tiruneh, and Baek]{kim2024sddgr}
Junsu Kim, Hoseong Cho, Jihyeon Kim, Yihalem~Yimolal Tiruneh, and Seungryul Baek.
\newblock Sddgr: Stable diffusion-based deep generative replay for class incremental object detection.
\newblock In \emph{CVPR}, 2024.

\bibitem[Kirkpatrick et~al.(2017)Kirkpatrick, Pascanu, Rabinowitz, Veness, Desjardins, Rusu, Milan, Quan, Ramalho, Grabska-Barwinska, et~al.]{kirkpatrick2017overcoming}
James Kirkpatrick, Razvan Pascanu, Neil Rabinowitz, Joel Veness, Guillaume Desjardins, Andrei~A Rusu, Kieran Milan, John Quan, Tiago Ramalho, Agnieszka Grabska-Barwinska, et~al.
\newblock Overcoming catastrophic forgetting in neural networks.
\newblock \emph{Proceedings of the National Academy of Sciences}, 2017.

\bibitem[Krause et~al.(2013)Krause, Stark, Deng, and Fei-Fei]{krause20133d}
Jonathan Krause, Michael Stark, Jia Deng, and Li Fei-Fei.
\newblock 3d object representations for fine-grained categorization.
\newblock In \emph{ICCVW}, 2013.

\bibitem[Krizhevsky et~al.(2009)Krizhevsky, Hinton, et~al.]{krizhevsky2009learning}
Alex Krizhevsky, Geoffrey Hinton, et~al.
\newblock Learning multiple layers of features from tiny images.
\newblock 2009.

\bibitem[Le and Yang(2015)]{le2015tiny}
Ya Le and Xuan Yang.
\newblock Tiny imagenet visual recognition challenge.
\newblock \emph{CS 231N}, 2015.

\bibitem[Li and Hoiem(2017)]{li2017learning}
Zhizhong Li and Derek Hoiem.
\newblock Learning without forgetting.
\newblock \emph{TPAMI}, 2017.

\bibitem[Liu et~al.(2020)Liu, Parisot, Slabaugh, Jia, Leonardis, and Tuytelaars]{liu2020more}
Yu Liu, Sarah Parisot, Gregory Slabaugh, Xu Jia, Ales Leonardis, and Tinne Tuytelaars.
\newblock More classifiers, less forgetting: A generic multi-classifier paradigm for incremental learning.
\newblock In \emph{ECCV}, 2020.

\bibitem[Loshchilov and Hutter(2016)]{loshchilov2016sgdr}
Ilya Loshchilov and Frank Hutter.
\newblock Sgdr: Stochastic gradient descent with warm restarts.
\newblock \emph{arXiv preprint arXiv:1608.03983}, 2016.

\bibitem[Loshchilov and Hutter(2017)]{loshchilov2017decoupled}
Ilya Loshchilov and Frank Hutter.
\newblock Decoupled weight decay regularization.
\newblock \emph{arXiv preprint arXiv:1711.05101}, 2017.

\bibitem[Maji et~al.(2013)Maji, Rahtu, Kannala, Blaschko, and Vedaldi]{maji2013fine}
Subhransu Maji, Esa Rahtu, Juho Kannala, Matthew Blaschko, and Andrea Vedaldi.
\newblock Fine-grained visual classification of aircraft.
\newblock \emph{arXiv preprint arXiv:1306.5151}, 2013.

\bibitem[McCloskey and Cohen(1989)]{mccloskey1989catastrophic}
Michael McCloskey and Neal~J Cohen.
\newblock Catastrophic interference in connectionist networks: The sequential learning problem.
\newblock In \emph{Psychology of learning and motivation}, pages 109--165. Elsevier, 1989.

\bibitem[Meng et~al.(2024)Meng, Zhang, Yang, Zhan, Zhao, and WAng]{meng2024diffclass}
Zichong Meng, Jie Zhang, Changdi Yang, Zheng Zhan, Pu Zhao, and Yanzhi WAng.
\newblock Diffclass: Diffusion-based class incremental learning.
\newblock In \emph{ECCV}, 2024.

\bibitem[Miller(1995)]{miller1995wordnet}
George~A Miller.
\newblock Wordnet: a lexical database for english.
\newblock \emph{Communications of the ACM}, 38\penalty0 (11):\penalty0 39--41, 1995.

\bibitem[M{\"u}ller et~al.(2019)M{\"u}ller, Kornblith, and Hinton]{muller2019does}
Rafael M{\"u}ller, Simon Kornblith, and Geoffrey~E Hinton.
\newblock When does label smoothing help?
\newblock \emph{NeurIPS}, 2019.

\bibitem[Ni et~al.(2023)Ni, Wei, Tang, Zhuang, and Tian]{ni2023continual}
Zixuan Ni, Longhui Wei, Siliang Tang, Yueting Zhuang, and Qi Tian.
\newblock Continual vision-language representation learning with off-diagonal information.
\newblock In \emph{ICML}, 2023.

\bibitem[Nilsback and Zisserman(2008)]{nilsback2008automated}
Maria-Elena Nilsback and Andrew Zisserman.
\newblock Automated flower classification over a large number of classes.
\newblock In \emph{2008 Sixth Indian Conference on Computer Vision, Graphics \& Image Processing}, 2008.

\bibitem[Parkhi et~al.(2012)Parkhi, Vedaldi, Zisserman, and Jawahar]{parkhi2012cats}
Omkar~M Parkhi, Andrea Vedaldi, Andrew Zisserman, and CV Jawahar.
\newblock Cats and dogs.
\newblock In \emph{CVPR}, 2012.

\bibitem[Pascanu and Bengio(2013)]{pascanu2013revisiting}
Razvan Pascanu and Yoshua Bengio.
\newblock Revisiting natural gradient for deep networks.
\newblock \emph{arXiv preprint arXiv:1301.3584}, 2013.

\bibitem[Radford et~al.(2021)Radford, Kim, Hallacy, Ramesh, Goh, Agarwal, Sastry, Askell, Mishkin, Clark, et~al.]{radford2021learning}
Alec Radford, Jong~Wook Kim, Chris Hallacy, Aditya Ramesh, Gabriel Goh, Sandhini Agarwal, Girish Sastry, Amanda Askell, Pamela Mishkin, Jack Clark, et~al.
\newblock Learning transferable visual models from natural language supervision.
\newblock In \emph{ICML}, 2021.

\bibitem[Rajasegaran et~al.(2019)Rajasegaran, Hayat, Khan, Khan, and Shao]{rajasegaran2019random}
Jathushan Rajasegaran, Munawar Hayat, Salman~H Khan, Fahad~Shahbaz Khan, and Ling Shao.
\newblock Random path selection for continual learning.
\newblock \emph{NeurIPS}, 2019.

\bibitem[Ramesh et~al.(2021)Ramesh, Pavlov, Goh, Gray, Voss, Radford, Chen, and Sutskever]{ramesh2021zero}
Aditya Ramesh, Mikhail Pavlov, Gabriel Goh, Scott Gray, Chelsea Voss, Alec Radford, Mark Chen, and Ilya Sutskever.
\newblock Zero-shot text-to-image generation.
\newblock In \emph{ICML}, 2021.

\bibitem[Rebuffi et~al.(2017)Rebuffi, Kolesnikov, Sperl, and Lampert]{rebuffi2017icarl}
Sylvestre-Alvise Rebuffi, Alexander Kolesnikov, Georg Sperl, and Christoph~H Lampert.
\newblock icarl: Incremental classifier and representation learning.
\newblock In \emph{CVPR}, 2017.

\bibitem[Rombach et~al.(2022)Rombach, Blattmann, Lorenz, Esser, and Ommer]{rombach2022high}
Robin Rombach, Andreas Blattmann, Dominik Lorenz, Patrick Esser, and Bj{\"o}rn Ommer.
\newblock High-resolution image synthesis with latent diffusion models.
\newblock In \emph{CVPR}, 2022.

\bibitem[Shi and Ye(2023)]{shi2023prototype}
Wuxuan Shi and Mang Ye.
\newblock Prototype reminiscence and augmented asymmetric knowledge aggregation for non-exemplar class-incremental learning.
\newblock In \emph{ICCV}, 2023.

\bibitem[Shi and Ye(2024)]{shiprospective}
Wuxuan Shi and Mang Ye.
\newblock Prospective representation learning for non-exemplar class-incremental learning.
\newblock \emph{NeurIPS}, 2024.

\bibitem[Shin et~al.(2017)Shin, Lee, Kim, and Kim]{shin2017continual}
Hanul Shin, Jung~Kwon Lee, Jaehong Kim, and Jiwon Kim.
\newblock Continual learning with deep generative replay.
\newblock \emph{NeurIPS}, 2017.

\bibitem[Smith et~al.(2021)Smith, Hsu, Balloch, Shen, Jin, and Kira]{smith2021always}
James Smith, Yen-Chang Hsu, Jonathan Balloch, Yilin Shen, Hongxia Jin, and Zsolt Kira.
\newblock Always be dreaming: A new approach for data-free class-incremental learning.
\newblock In \emph{ICCV}, 2021.

\bibitem[Tian et~al.(2023)Tian, He, Dai, Ma, Liu, and Kira]{tian2023trainable}
Junjiao Tian, Zecheng He, Xiaoliang Dai, Chih-Yao Ma, Yen-Cheng Liu, and Zsolt Kira.
\newblock Trainable projected gradient method for robust fine-tuning.
\newblock In \emph{CVPR}, 2023.

\bibitem[Tian et~al.(2024)Tian, Fan, Isola, Chang, and Krishnan]{tian2024stablerep}
Yonglong Tian, Lijie Fan, Phillip Isola, Huiwen Chang, and Dilip Krishnan.
\newblock Stablerep: Synthetic images from text-to-image models make strong visual representation learners.
\newblock \emph{NeurIPS}, 2024.

\bibitem[Van~de Ven and Tolias(2019)]{van2019three}
Gido~M Van~de Ven and Andreas~S Tolias.
\newblock Three scenarios for continual learning.
\newblock \emph{arXiv preprint arXiv:1904.07734}, 2019.

\bibitem[Wang et~al.(2022{\natexlab{a}})Wang, Zhou, Ye, and Zhan]{wang2022foster}
Fu-Yun Wang, Da-Wei Zhou, Han-Jia Ye, and De-Chuan Zhan.
\newblock Foster: Feature boosting and compression for class-incremental learning.
\newblock In \emph{ECCV}, 2022{\natexlab{a}}.

\bibitem[Wang et~al.(2022{\natexlab{b}})Wang, Huang, and Hong]{wang2022s}
Yabin Wang, Zhiwu Huang, and Xiaopeng Hong.
\newblock S-prompts learning with pre-trained transformers: An occam’s razor for domain incremental learning.
\newblock \emph{NeurIPS}, 2022{\natexlab{b}}.

\bibitem[Wang et~al.(2022{\natexlab{c}})Wang, Zhang, Ebrahimi, Sun, Zhang, Lee, Ren, Su, Perot, Dy, et~al.]{wang2022dualprompt}
Zifeng Wang, Zizhao Zhang, Sayna Ebrahimi, Ruoxi Sun, Han Zhang, Chen-Yu Lee, Xiaoqi Ren, Guolong Su, Vincent Perot, Jennifer Dy, et~al.
\newblock Dualprompt: Complementary prompting for rehearsal-free continual learning.
\newblock In \emph{ECCV}, 2022{\natexlab{c}}.

\bibitem[Wang et~al.(2022{\natexlab{d}})Wang, Zhang, Lee, Zhang, Sun, Ren, Su, Perot, Dy, and Pfister]{wang2022learning}
Zifeng Wang, Zizhao Zhang, Chen-Yu Lee, Han Zhang, Ruoxi Sun, Xiaoqi Ren, Guolong Su, Vincent Perot, Jennifer Dy, and Tomas Pfister.
\newblock Learning to prompt for continual learning.
\newblock In \emph{CVPR}, 2022{\natexlab{d}}.

\bibitem[Wortsman et~al.(2022)Wortsman, Ilharco, Kim, Li, Kornblith, Roelofs, Lopes, Hajishirzi, Farhadi, Namkoong, et~al.]{wortsman2022robust}
Mitchell Wortsman, Gabriel Ilharco, Jong~Wook Kim, Mike Li, Simon Kornblith, Rebecca Roelofs, Raphael~Gontijo Lopes, Hannaneh Hajishirzi, Ali Farhadi, Hongseok Namkoong, et~al.
\newblock Robust fine-tuning of zero-shot models.
\newblock In \emph{CVPR}, 2022.

\bibitem[Wu et~al.(2019)Wu, Chen, Wang, Ye, Liu, Guo, and Fu]{wu2019large}
Yue Wu, Yinpeng Chen, Lijuan Wang, Yuancheng Ye, Zicheng Liu, Yandong Guo, and Yun Fu.
\newblock Large scale incremental learning.
\newblock In \emph{CVPR}, 2019.

\bibitem[Xiang et~al.(2019)Xiang, Fu, Ji, and Huang]{xiang2019incremental}
Ye Xiang, Ying Fu, Pan Ji, and Hua Huang.
\newblock Incremental learning using conditional adversarial networks.
\newblock In \emph{ICCV}, 2019.

\bibitem[Xiao et~al.(2010)Xiao, Hays, Ehinger, Oliva, and Torralba]{xiao2010sun}
Jianxiong Xiao, James Hays, Krista~A Ehinger, Aude Oliva, and Antonio Torralba.
\newblock Sun database: Large-scale scene recognition from abbey to zoo.
\newblock In \emph{IEEE Computer Society Conference on Computer Vision and Pattern Recognition}, 2010.

\bibitem[Yan et~al.(2021)Yan, Xie, and He]{yan2021dynamically}
Shipeng Yan, Jiangwei Xie, and Xuming He.
\newblock Der: Dynamically expandable representation for class incremental learning.
\newblock In \emph{CVPR}, 2021.

\bibitem[Ye et~al.(2025)Ye, Rong, Huang, Du, Yu, and Tao]{ye2025survey}
Mang Ye, Xuankun Rong, Wenke Huang, Bo Du, Nenghai Yu, and Dacheng Tao.
\newblock A survey of safety on large vision-language models: Attacks, defenses and evaluations.
\newblock \emph{arXiv preprint arXiv:2502.14881}, 2025.

\bibitem[Yin et~al.(2020)Yin, Molchanov, Alvarez, Li, Mallya, Hoiem, Jha, and Kautz]{yin2020dreaming}
Hongxu Yin, Pavlo Molchanov, Jose~M Alvarez, Zhizhong Li, Arun Mallya, Derek Hoiem, Niraj~K Jha, and Jan Kautz.
\newblock Dreaming to distill: Data-free knowledge transfer via deepinversion.
\newblock In \emph{CVPR}, 2020.

\bibitem[Yoon et~al.(2017)Yoon, Yang, Lee, and Hwang]{yoon2017lifelong}
Jaehong Yoon, Eunho Yang, Jeongtae Lee, and Sung~Ju Hwang.
\newblock Lifelong learning with dynamically expandable networks.
\newblock \emph{arXiv preprint arXiv:1708.01547}, 2017.

\bibitem[Yu et~al.(2024)Yu, Zhuge, Zhang, Hu, Wang, Lu, and He]{yu2024boosting}
Jiazuo Yu, Yunzhi Zhuge, Lu Zhang, Ping Hu, Dong Wang, Huchuan Lu, and You He.
\newblock Boosting continual learning of vision-language models via mixture-of-experts adapters.
\newblock In \emph{CVPR}, 2024.

\bibitem[Zenke et~al.(2017)Zenke, Poole, and Ganguli]{zenke2017continual}
Friedemann Zenke, Ben Poole, and Surya Ganguli.
\newblock Continual learning through synaptic intelligence.
\newblock In \emph{ICML}, 2017.

\bibitem[Zheng et~al.(2023)Zheng, Ma, Wang, Qin, Yue, and You]{zheng2023preventing}
Zangwei Zheng, Mingyuan Ma, Kai Wang, Ziheng Qin, Xiangyu Yue, and Yang You.
\newblock Preventing zero-shot transfer degradation in continual learning of vision-language models.
\newblock In \emph{ICCV}, 2023.

\bibitem[Zhou et~al.(2023)Zhou, Zhang, Ning, Ye, Zhan, and Liu]{zhou2023learning}
Da-Wei Zhou, Yuanhan Zhang, Jingyi Ning, Han-Jia Ye, De-Chuan Zhan, and Ziwei Liu.
\newblock Learning without forgetting for vision-language models.
\newblock \emph{arXiv preprint arXiv:2305.19270}, 2023.

\bibitem[Zhou et~al.(2022{\natexlab{a}})Zhou, Yang, Loy, and Liu]{zhou2022conditional}
Kaiyang Zhou, Jingkang Yang, Chen~Change Loy, and Ziwei Liu.
\newblock Conditional prompt learning for vision-language models.
\newblock In \emph{CVPR}, 2022{\natexlab{a}}.

\bibitem[Zhou et~al.(2022{\natexlab{b}})Zhou, Yang, Loy, and Liu]{zhou2022learning}
Kaiyang Zhou, Jingkang Yang, Chen~Change Loy, and Ziwei Liu.
\newblock Learning to prompt for vision-language models.
\newblock \emph{IJCV}, 2022{\natexlab{b}}.

\bibitem[Zhu et~al.(2021)Zhu, Zhang, Wang, Yin, and Liu]{zhu2021prototype}
Fei Zhu, Xu-Yao Zhang, Chuang Wang, Fei Yin, and Cheng-Lin Liu.
\newblock Prototype augmentation and self-supervision for incremental learning.
\newblock In \emph{CVPR}, 2021.

\end{thebibliography}
